\documentclass{article}

\PassOptionsToPackage{numbers, compress}{natbib}

\usepackage{hyperref}       

\usepackage{amsmath,amsfonts,bm}

\def\eqref#1{equation~\ref{#1}}

\def\1{\bm{1}}

\DeclareMathAlphabet{\mathsfit}{\encodingdefault}{\sfdefault}{m}{sl}
\SetMathAlphabet{\mathsfit}{bold}{\encodingdefault}{\sfdefault}{bx}{n}

\DeclareMathOperator*{\argmax}{arg\,max}

\newcommand{\mytitle}{\alg: Stabilizing Dot-regression with Global Feature Distillation for Federated Learning}

\newcommand{\alg}{\textbf{\code{FedDr+}}\xspace}

\usepackage{geometry}
\usepackage{booktabs} 
\usepackage{bbm}
\usepackage{mathtools}
\usepackage{nccmath}
\usepackage{setspace}
\usepackage[T1]{fontenc}
\usepackage[scaled]{beramono}
\usepackage{subcaption}

\usepackage{algorithm} 
\usepackage{algorithmic}  
\usepackage[linesnumbered,ruled,vlined,algo2e]{algorithm2e}
\SetKwInput{KwInput}{Input}                
\SetKwInput{KwOutput}{Output}              

\SetCommentSty{mycommfont}

\SetAlCapSty{algcapsty}

\usepackage[T1]{fontenc}
\usepackage{wrapfig,lipsum,booktabs}

\usepackage{soul}
\usepackage{dsfont}
\usepackage{enumerate}
\usepackage{enumitem}

\usepackage{amsmath}
\usepackage{amsfonts}
\usepackage{bbm}
\usepackage{dsfont}
\usepackage[Symbol]{upgreek}
\usepackage{lscape}
\usepackage{caption}
\usepackage{balance}
\usepackage{xspace}
\usepackage{float}

\usepackage{wasysym}
\usepackage[table,xcdraw,dvipsnames]{xcolor}
\usepackage{multirow}
\usepackage{array, boldline, rotating}

\usepackage{amssymb}
\usepackage{pifont}

\newcommand{\mc}[1]{\mathcal{#1}}

\definecolor{LightCyan}{rgb}{0.88,1,1}
\definecolor{Blue}{rgb}{0, 0.5, 1}
\definecolor{Green}{rgb}{0.0, 0.8, 0.0 }
\definecolor{Red}{rgb}{0.95, 0.55, 0.6}
\definecolor{Skyblue}{rgb}{0.6, 0.6, 0.95 }

\renewcommand*\eqref[1]{(\ref{#1})}

\newcommand{\sungnyun}[1]{{\color{RoyalBlue}$\rightarrow$SK: #1}}
\newcommand{\eg}{\emph{e.g.,~}}
\newcommand{\ie}{\emph{i.e.,~}}

\newcommand{\myparagraph}[1]{\vspace{0.07cm}\noindent\textbf{#1}~}

\def\code#1{\texttt{#1}}

\NewDocumentCommand{\supptitle}{s}{
\onecolumn
\begin{center}
    \rule{\textwidth}{0.03cm}\\[0.1cm]
    - Appendix -\\[0.2cm]
    {\Large 
        \textbf{\mytitle }
    }\\
    \rule{\textwidth}{0.03cm}\\[0.2cm]
\end{center}
}

\usepackage[preprint]{neurips_2024}

\usepackage[utf8]{inputenc} 
\usepackage[T1]{fontenc}    
\usepackage{url}            
\usepackage{booktabs}       
\usepackage{amsfonts}       
\usepackage{nicefrac}       
\usepackage{xcolor}         

\usepackage{graphicx}
\usepackage{mathtools}
\usepackage{booktabs}

\usepackage{multirow}
\usepackage{adjustbox}
\usepackage{wrapfig}
\usepackage{kotex}
\usepackage{enumitem}
\usepackage{amsmath}
\usepackage{lipsum}
\usepackage{blindtext}

\usepackage{amsthm}
\usepackage{amsfonts}
\usepackage{dsfont}
\usepackage{xcolor}
\usepackage{subcaption}
\usepackage{booktabs}
\usepackage{hyperref}

\definecolor{darkblue}{rgb}{0, 0, 0.5}

\hypersetup{
  colorlinks   = true, 
  urlcolor     = darkblue, 
  linkcolor    = darkblue, 
  citecolor   = cyan 
}
\theoremstyle{plain}
\newtheorem{thm}{Theorem}[section]
\newtheorem{cor}{Corollary}
\newtheorem{lem}{Lemma}
\newtheorem{prop}{Proposition}

\theoremstyle{definition}
\newtheorem{defn}{Definition}

\usepackage[capitalize,noabbrev]{cleveref}
\crefname{section}{Sec.}{Secs.}
\Crefname{section}{Section}{Sections}
\Crefname{table}{Table}{Tables}
\crefname{table}{Tab.}{Tabs.}

\def\equationautorefname~#1\null{Eq.~(#1)\null}

\title{\mytitle}

\author{
  Seongyoon Kim\\
  Dept. ISysE, KAIST\\
  \texttt{curisam@kaist.ac.kr}\\
  \And
  Minchan Jeong\\
  Graduate School of AI, KAIST\\
  \texttt{mcjeong@kaist.ac.kr}\\
  \And
  Sungnyun Kim\\
  Graduate School of AI, KAIST\\
  \texttt{ksn4397@kaist.ac.kr}\\
  \And
  Sungwoo Cho\\
  Graduate School of AI, KAIST\\
  \texttt{peter8526@kaist.ac.kr}\\
  \And
  Sumyeong Ahn\thanks{corresponding authors}\\
  CSE, Michigan State University\\
  \texttt{sumyeong@msu.edu}\\
  \And
  {Se-Young Yun}\footnotemark[1]\\
  Graduate School of AI, KAIST\\
  \texttt{yunseyoung@gmail.com}\\
}

\begin{document}

\maketitle

\begin{abstract}
Federated Learning (FL) has emerged as a pivotal framework for the development of effective global models (global FL) or personalized models (personalized FL) across clients with heterogeneous, non-iid data distribution. A key challenge in FL is client drift, where data heterogeneity impedes the aggregation of scattered knowledge. Recent studies have tackled the client drift issue by identifying significant divergence in the last classifier layer. To mitigate this divergence, strategies such as freezing the classifier weights and aligning the feature extractor accordingly have proven effective. Although the local alignment between classifier and feature extractor has been studied as a crucial factor in FL, we observe that it may lead the model to overemphasize the observed classes within each client. Thus, our objectives are twofold: (1) enhancing local alignment while (2) preserving the representation of unseen class samples. This approach aims to effectively integrate knowledge from individual clients, thereby improving performance for both global and personalized FL. To achieve this, we introduce a novel algorithm named \alg, which empowers local model alignment using dot-regression loss. \alg freezes the classifier as a simplex ETF to align the features and improves aggregated global models by employing a feature distillation mechanism to retain information about unseen/missing classes. Consequently, we provide empirical evidence demonstrating that our algorithm surpasses existing methods that use a frozen classifier to boost alignment across the diverse distribution.
\end{abstract}

\vspace{-20pt}
\section{Introduction}
\label{sec:intro}

Federated Learning (FL)~\cite{mcmahan2017communication, oh2021fedbabu, he2020fedml} is a privacy-aware distributed learning strategy that employs data from multiple clients while ensuring their data privacy. A foundational method in FL, known as {FedAvg}~\cite{mcmahan2017communication}, involves four iterative phases: (1) distributing a global model to clients, (2) training local models using each client's private dataset, (3) transmitting the locally trained models back to the server, and (4) aggregating these models. This method effectively protects privacy without requiring the transmission of raw data to the server. However, a significant challenge in FL is data heterogeneity, called \emph{non-iidness}, which refers to the different underlying data distribution across clients. Such variance can cause \emph{client drift} during training, obstructing the convergence of the aggregated model and significantly reducing its effectiveness.

To address client drift in non-iid scenarios, recent works~\cite{oh2021fedbabu, dong2022spherefed, li2023no,  fan2024federated} have identified that the last classifier layer in neural networks is particularly vulnerable to this issue. Hence, they suggest strategies that freeze the classifier while updating only the feature extractor. These approaches aim to enhance the \emph{alignment} between the frozen classifier and the output from the feature extractor. For instance, {FedBABU}~\cite{oh2021fedbabu} employs various classifier initialization techniques, keeping it fixed during the training of the feature extractor. The methods proposed in~\cite{dong2022spherefed, li2023no, fan2024federated, huang2023neural, xiao2024fedloge} utilize more robust initialization, the Equiangular Tight Frame (ETF) classifier~\cite{papyan2020prevalence}, to replace traditional random initialization and improve the local alignment strategy.

\begin{wrapfigure}[24]{r}{0.62\linewidth}
  \centering
  \small
  \vspace{-13pt}
  \includegraphics[width=\linewidth]{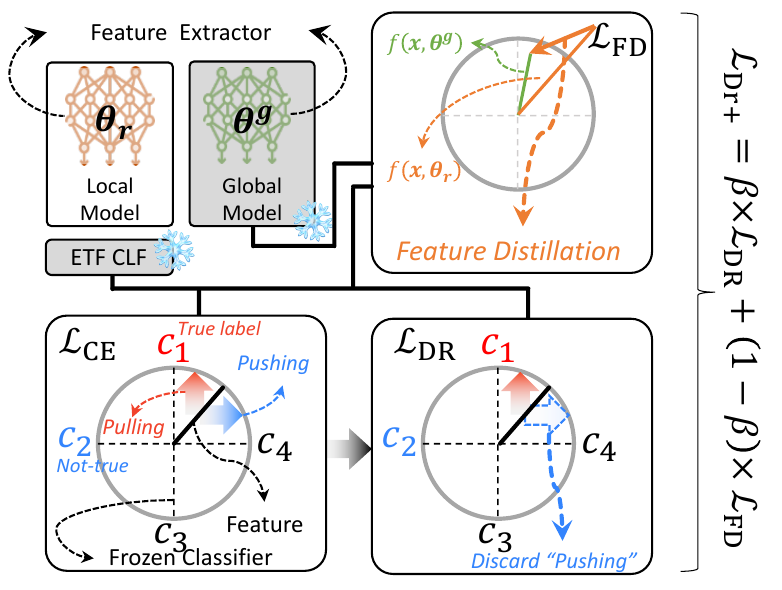}
  \vspace{-6pt}
  \caption{Overview of the proposed method, \alg trained with $\mc{L}_{\text{Dr+}}$. To enhance the local alignment, we employ dot-regression loss $\mc{L}_{\text{DR}}$, which discards the pushing term of cross-entropy loss, and propose a feature distillation $\mc{L}_{\text{FD}}$ to preserve the knowledge imbued in the global model. We describe $\mc{L}_{\text{DR}}$ in~\autoref{sec:pre}, and feature distillation in~\autoref{sec:prob} in detail.}
  \vspace{45pt}
  \label{fig:overview}
\end{wrapfigure}

A frozen classifier is also extensively explored in other research areas, such as class imbalance~\cite{yang2022inducing} and class incremental learning~\cite{yang2023neural}, with a consistent objective similar to aforementioned FL studies---enhancing alignment. Recently, these fields have advanced by introducing and utilizing a novel type of loss, called dot-regression loss $\mc{L}_\text{DR}$, which aims to achieve alignment rapidly. In summary, $\mc{L}_\text{DR}$ originates from the decomposition analysis of cross-entropy (CE) loss, which includes \emph{pulling} and \emph{pushing} components.
As suggested in~\cite{yang2022inducing}, the \emph{pulling} component is a force that {attracts} features to the target class, whereas the \emph{pushing} component is a force that drives features {away from} other non-target classes.
$\mc{L}_\text{DR}$ discards the \emph{pushing} component, as it slows down convergence to the desired alignment (refer to~\autoref{fig:overview}).

Following the advancement of leveraging the frozen classifier with dot-regression loss, we investigate the application of this loss to FL. 
However, our findings indicate that dot-regression loss does not necessarily lead to sufficient performance improvement of the aggregated server-side model, although it enhances \emph{local alignment} as intended. 
We observe that this drawback stems from the handling of unseen class samples. Specifically, while alignment improves for the classes in the local training dataset, it significantly deteriorates for unseen classes. This observation highlights the need to preserve the representation of unobserved classes during local training. To address this issue, we propose a training mechanism, termed \alg, that employs dot-regression loss alongside feature distillation that reduces the distance between feature vectors of local and global models.
\vspace{7pt}

\myparagraph{Contributions.} Our main contributions are summarized as follows:
\vspace{-5pt}
\begin{itemize}[leftmargin=15pt]
    \item We find that dot-regression loss is not easily compatible with FL, although it can enhance the alignment of seen classes. The drawback comes from a significant loss of information on unseen classes, which is vital in the global model perspective. 
    Therefore, we aim to preserve information of unseen classes within the FL system.
    \item To preserve global knowledge, including unseen class information while maintaining the advantages of $\mc{L}_\text{DR}$, we propose \alg, which utilizes a feature distillation when training local models. This regularizer prevents the model from focusing solely on the local alignment.
    \item We verify that the proposed method surpasses the conventional algorithms in both global and personalized FL under various datasets and non-iid settings. 
\end{itemize}
\section{Preliminaries}
\label{sec:pre}


\subsection{Basic Setup of Conventional FedAvg Pipeline}
\label{subsec:fl}

\myparagraph{Basic FL setup.}
Let $[N] = \{1, \ldots, N\}$ denote the indices of clients, each with a unique training dataset $D_{\text{train}}^{i} = \{(x_m, y_m)\}_{m=1}^{|D_{\text{train}}^{i}|}$, where $(x_m, y_m) \sim \mc{D}^{i}$ for the $i^{\text{th}}$ client, $x_m$ is the input data, and $y_m \in [C]$ is the corresponding label among $C$ classes. Importantly, FL studies predominantly address the scenario where the data distributions are heterogeneous, \ie $\mc{D}^{i}$ varies across clients. Knowledge distributed among clients is collected over $R$ communication rounds.
The general objective of FL is to train a model fit to the aggregated knowledge, $\bigcup_{i \in [N]} \mc{D}^i$. This objective can be seen as solving the optimization problem:
\begin{equation}
\label{eq:flgoal}
\min_{\bm{\Theta} = (\bm{\theta}, \bm{V})} \sum_{i \in [N]} \frac{|D_{\text{train}}^{i}|}{\sum_{j \in [N]} |D_{\text{train}}^{j}|} \, \scalebox{1.45}{$\displaystyle\mathop{\mathbb{E}}$}_{(x, y) \sim \mc{D}^i} \Big[\mc{L}(x, y; \bm{\theta}, \bm{V})\Big]\,,
\end{equation}
where $\mc{L}$ is the instance-wise loss function, $\bm{\theta}$ is the weight parameter for the feature extractor, and $\bm{V} = [v_1, \ldots, v_C]\in \mathbb{R}^{d \times C}$ is the classifier weight matrix. We use the notation $\bm{\Theta}$ to denote the entire set of model parameters.

At the beginning of each round $r \in [R]$, the server has access to only a subset of clients $\mc{S}_{r} \subset [N]$ participating in the $r^\text{th}$ round. At each round $r$, the server transmits the global model parameters $\bm{\Theta}_{r-1}^g$ to the participating clients. Each client then updates the parameters with their private data $D_{\text{train}}^{i}$ and uploads $\bm{\Theta}_{r}^i$ to the global server. By incorporating the locally trained weights, the server then updates the global model parameters to $\bm{\Theta}_{r}^g$.

\myparagraph{{FedAvg} pipeline.}
Our study follows the conventional {FedAvg}~\cite{mcmahan2017communication} framework to address the FL problem. {FedAvg} updates the global model parameters from locally trained parameters by aggregating these local models into $\bm{\Theta}_{r}^{g} = \sum_{i \in S_r} w_r^i \bm{\Theta}_{r}^{i}$, where $w_r^i = |D_{\text{train}}^{i}| \,/\, {\sum_{j \in S_{r}} |D_{\text{train}}^{j}|}$ is the importance weight of the $i^{\text{th}}$ client. 


\subsection{Dot-Regression Loss for Faster Feature Alignment}
\myparagraph{Dot-regression loss $\mc{L}_{\text{DR}}$.}
This loss~\cite{yang2022inducing}
facilitates a faster alignment of feature vectors (penultimate layer
outputs) $f(x;\bm{\theta})\in \mathbb{R}^{d}$ to the true class direction of $v_y$, reducing the cosine angle as follows:
\begin{equation}
    \mc{L}_{\text{DR}} (x, y; \bm{\theta}, \bm{V}) = \frac{1}{2}\Big(\cos\big(f(x; \bm{\theta}), v_y\big) - 1\Big)^2
\end{equation}
where $\cos(\mathrm{vec}_1, \mathrm{vec}_2)$ denotes the cosine of the angle between two vectors  $\angle (\mathrm{vec}_1,\mathrm{vec}_2)$. 

The main motivation is that the gradient of the cross-entropy (CE) loss for the feature vector can be decomposed into a \emph{pulling} and \emph{pushing} gradient, and recent work indicates that we can achieve better convergence by removing the pushing effect~\cite{yang2022inducing, li2021fedrs}. The \emph{pulling} gradient aligns $f(x; \bm{\theta})$ with $v_y$, while the \emph{pushing} gradient ensures $f(x; \bm{\theta})$ does not align with $v_c$ for all $c \neq y$ (\autoref{app:prelim} details the exact form of pulling and pushing gradients).
Since $\mc{L}_{\text{DR}}$ directly attracts features to the true-class classifier, it drops the \emph{pushing} gradient, thereby increasing the convergence speed for maximizing $\cos(f(x; \bm{\theta}), v_y)$. 

\myparagraph{Frozen ETF classifier.}
Since $\mc{L}_{\text{DR}}$ focuses on aligning feature vectors with the true-class classifier, the classifier is not required to be trained. Instead, we construct the classifier to satisfy the simplex Equiangular Tight Frame (ETF) condition, a constructive way to achieve maximum angular separation between class vectors~\cite{yang2022inducing, yang2023neural}. Concretely, we initialize the classifier weight $\bm{V}$ as follows and freeze it throughout training: 
\begin{equation}
\label{eq:etf}
\bm{V} \longleftarrow \sqrt{\frac{C}{C-1}} \bm{U} \left( \bm{I}_C - \frac{1}{C} \bm{1}_C \bm{1}_C^\top \right),
\end{equation}
where $\bm{U} \in \mathbb{R}^{d \times C}$ is a randomly initialized orthogonal matrix. Note that each $v_i$ in the classifier weight $\bm{V}$ satisfies $\cos(v_i, v_j) = -\frac{1}{C-1}$ for all $i \neq j \in [C]$\footnote{This relation for cosines holds if the $v_i$'s are symmetrically distributed such that $\bar{v} = \frac{1}{C} \sum_{i \in [C]} v_i = 0$, and $\cos(v_i, v_j)$ are all the same for $i \neq j$ .}.

\section{When Dot-Regression Loss Meets Federated Learning}
\label{sec:prob}

Given our focus on applying $\mathcal{L}_{\text{DR}}$ to FL, we first examine its impact on FL models compared to the CE loss $\mathcal{L}_{\text{CE}}$. In summary, we find that while $\mathcal{L}_{\text{DR}}$ improves alignment and performance on \textcolor{red}{observed} class labels, it faces challenge with \textcolor{blue}{unobserved} classes\footnote{While we use the term ``unobserved'' in this context, it also applies to ``rarely'' existing classes.}, which are essential for the generalization objective. To address this issue, we propose \alg, which integrates $\mathcal{L}_{\text{DR}}$ with a novel feature distillation loss. We then evaluate \alg by analyzing the effect of feature distillation and compare it with various FL algorithms and regularizers.

\myparagraph{Experimental configuration.} 
In this section, we conduct experiments on CIFAR-100~\cite{krizhevsky2009cifar} with a shard non-iid setting ($s$=10), where each client contains at most 10 classes. We additionally employ LDA setting ($\alpha$=0.1) in \autoref{subsec:synergy_effect}. Refer to \autoref{sec:exp} for more details on the dataset configuration. The model is trained for 320 communication rounds, randomly selecting 10\% of clients in each round, and the learning rate is decayed at $160^\text{th}$ and $240^\text{th}$ rounds.

\begin{figure}[t]
    \centering
    \includegraphics[width=0.35\textwidth]{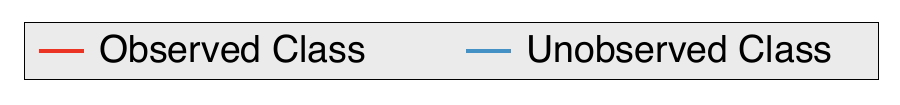}\\
    \begin{minipage}{0.48\textwidth}
        \centering
        \begin{subfigure}[b]{0.45\textwidth}
            \centering
            \includegraphics[width=1\textwidth]{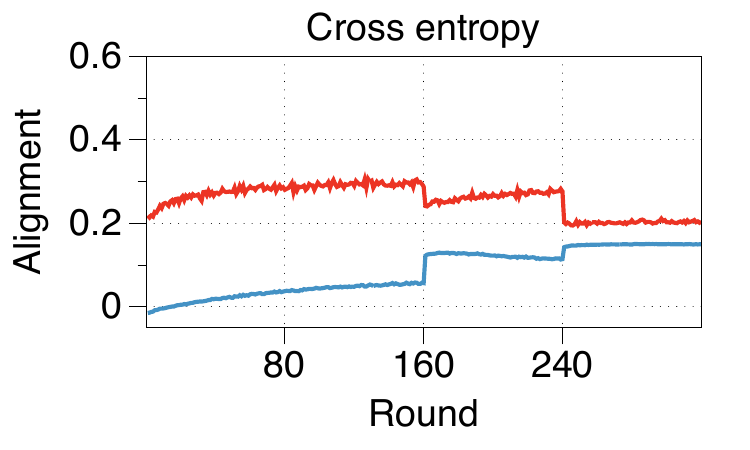}
            \includegraphics[width=1\textwidth]{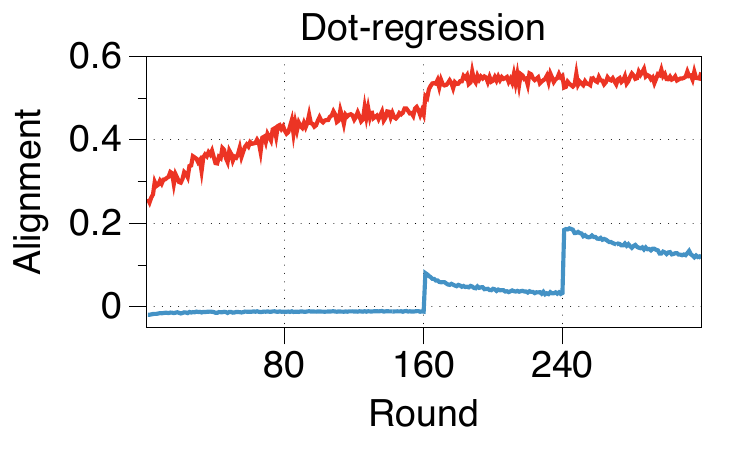}
            \vspace{-15pt}
            \subcaption{Local alignment}
        \end{subfigure}
        \begin{subfigure}[b]{0.45\textwidth}
            \centering
            \includegraphics[width=1\textwidth]{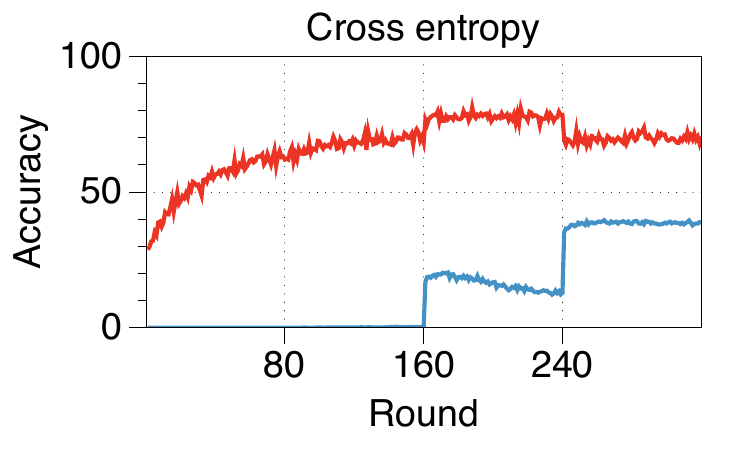}
            \includegraphics[width=1\textwidth]{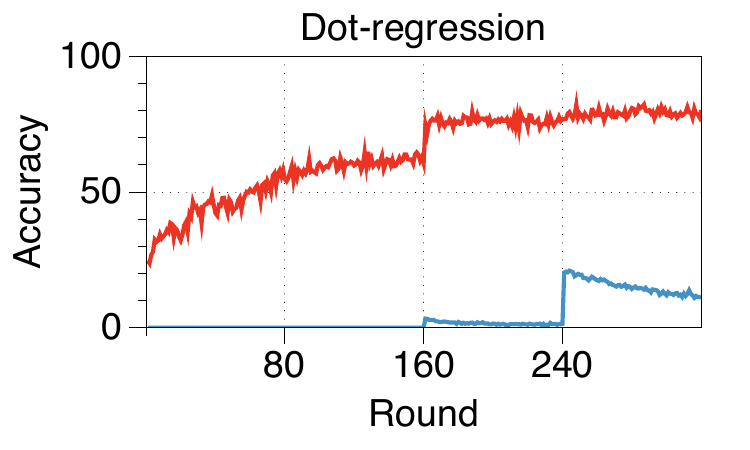}
            \vspace{-15pt}
            \subcaption{Local accuracy}
        \end{subfigure}
        \caption{Comparison of (a) feature-classifier alignment and (b) accuracy on the \textcolor{red}{observed} and \textcolor{blue}{unobserved} classes test data for $\bm{\theta}_r^i$ trained with $\mathcal{L}_\text{CE}$ and $\mathcal{L}_\text{DR}$.}
        \vspace{-10pt}
        \label{fig:local_align_acc}
    \end{minipage}
    \hfill
    \begin{minipage}{0.48\textwidth}
        \centering
        \begin{subfigure}[b]{0.45\textwidth}
            \centering
            \includegraphics[width=1\textwidth]{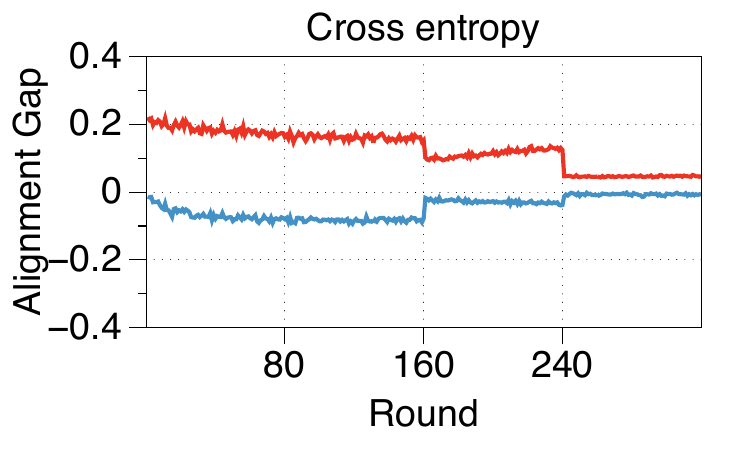}
            \includegraphics[width=1\textwidth]{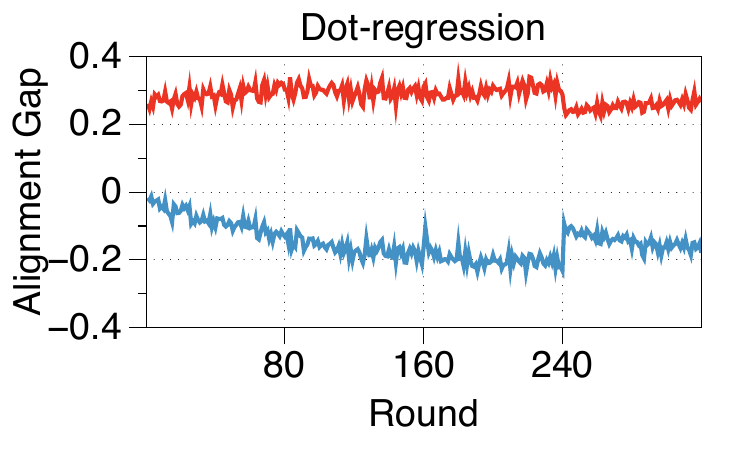}
            \vspace{-15pt}
            \subcaption{Alignment gap}
        \end{subfigure}
        \begin{subfigure}[b]{0.45\textwidth}
            \centering
            \includegraphics[width=1\textwidth]{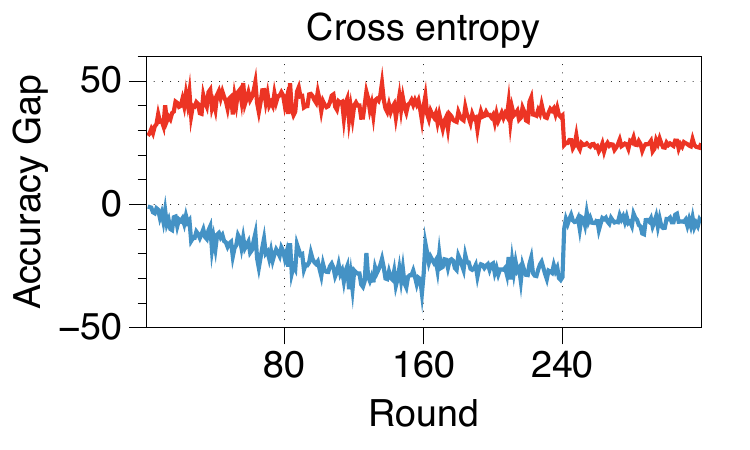}
            \includegraphics[width=1\textwidth]{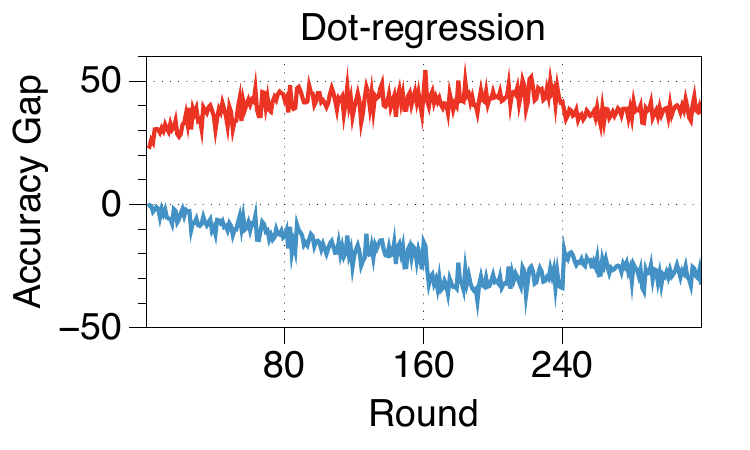}
            \vspace{-15pt}
            \subcaption{Accuracy gap}
        \end{subfigure}
        \caption{Comparison of (a) feature-classifier alignment gap  and (b) accuracy gap on the \textcolor{red}{observed} and \textcolor{blue}{unobserved} classes test data for $\bm{\theta}_r^i$ trained with $\mathcal{L}_\text{CE}$ and $\mathcal{L}_\text{DR}$.}
        \vspace{-5pt}
        \label{fig:gain_align_acc}
    \end{minipage}
\end{figure}

\subsection{Impact of Dot-Regression Loss on Local and Global Models}
\label{subsec:dotreg}

We investigate the performance of local models on average when trained with $\mc{L}_{\text{DR}}$ compared to $\mc{L}_{\text{CE}}$. In~\autoref{fig:local_align_acc}--\ref{fig:gain_align_acc}, we calculate the statistics on two datasets: the {\textcolor{red}{observed} class set} $\mc{O}^i$, which includes classes present in each client's training data $D_\text{train}^i$, and the {\textcolor{blue}{unobserved} class set} $\mc{U}^i$, consisting of classes unseen during training. This partition highlights the challenges associated with generalizing to unseen classes in FL.

First, we evaluate the feature-classifier alignment $\cos(f(x; \bm{\theta}_r^i), v_y)$ and accuracy of each local model on the test data 
(\autoref{fig:local_align_acc}). 
We then observe the amount of change from the given global model to each local model in every communication round (\autoref{fig:gain_align_acc}). For instance, the alignment gap is denoted by $\cos(f(x;\bm{\theta}_r^i), v_y) - \cos(f(x;\bm{\theta}_{r-1}^g),v_y)$.



\myparagraph{Performance analysis of local models.} \autoref{fig:local_align_acc} shows that $\mc{L}_\text{DR}$, by focusing its pulling effects exclusively on \textcolor{red}{observed} classes within a client's dataset, effectively \textcolor{red}{\textit{enhances alignment and accuracy}} for these classes. However, this specificity leads to \textcolor{blue}{\textit{poor generalization}} on \textcolor{blue}{unobserved} classes, resulting in significantly weaker performance than models trained with $\mathcal{L}_\text{CE}$. \autoref{fig:gain_align_acc} displays the different impacts on $\mc{O}^i$ and $\mc{U}^i$ during updates from the global model to local models. $\mathcal{L}_\text{DR}$ significantly boosts alignment and accuracy for $\mc{O}^i$ but causes significant reductions for $\mc{U}^i$.

\myparagraph{Global model accuracy result.}
We confirm that $\mathcal{L}_\text{DR}$ shows superior accuracy for $\mc{O}^i$ compared to $\mathcal{L}_\text{CE}$ but is less effective at generalizing to $\mc{U}^i$. In the shard setting ($s=10$)—where each client has access to at most 10 out of 100 classes—this shortcoming significantly reduces the global model's overall accuracy ($\mathcal{L}_\text{DR}$: 42.52$\%$ vs. $\mathcal{L}_\text{CE}$: 46.38$\%$). Thus, it is crucial to develop methods that retain the strengths of $\mathcal{L}_\text{DR}$, \ie alignment of \textcolor{red}{observed} classes, while improving generalization for \textcolor{blue}{unobserved} classes, highlighting the need for more adaptive loss functions in FL.

\subsection{\alg: Dot-Regression and Feature Distillation for Federated Learning}\label{subsec:method}

We propose \alg to mitigate forgetting unobserved classes while retaining the strengths of dot-regression loss in aligning features of observed classes. Using $\mathcal{L}_\text{DR}$ with the frozen classifier $\bm{V}$, \alg includes a regularizer that fully distills the global model's feature vectors $f(x;\bm{\theta}^g)\in\mathbb{R}^{d}$ to the client features $f(x;\bm{\theta})$, to enhance generalization across all classes. The proposed loss function $\mc{L}_{\text{Dr+}}$, shown in \autoref{alg:FedDR}, combines $\mc{L}_\text{DR}$ with a regularizer $\mc{L}_\text{FD}(x; \bm{\theta}, \bm{\theta}^{g})=\frac{1}{d}\|f(x;\bm{\theta})-f(x;\bm{\theta}^g)\|_2^2$. Unless specified, we use a scaling parameter $\beta = 0.9$ throughout the paper.
\begin{equation}\label{alg:FedDR}
    \mc{L}_{\text{Dr+}}(x,y; \bm{\theta}, \bm{\theta}^g, \bm{V})=\beta \cdot \mc{L}_{\text{DR}}(x,y;\bm{\theta}, \bm{V})+(1-\beta)\cdot \mc{L}_{\text{FD}}(x; \bm{\theta}, \bm{\theta}^{g})
\end{equation}

\myparagraph{Why feature distillation?}
To address data heterogeneity in FL, various distillation methods have been explored, including model parameters~\cite{oh2021fedbabu, li2020federated, he2020group, li2019fedmd}, logit-related measurement~\cite{li2019fedmd, lee2022preservation, itahara2021distillation, ye2023fake, lin2020ensemble, chen2019knowledge, qian2022switchable}, and co-distillation~\cite{chen2024spectral, cho2023communication}. 
In contrast, we utilize the \textit{feature} distillation~\cite{heo2019comprehensive} technique because the feature directly concerns alignment. On the other hand, logits lose information from features when projected onto a frozen ETF classifier~\cite{heo2019comprehensive, li2017mimicking, li2023rethinking, ben2022s}.
By distilling features, we leverage the global, differentiated knowledge for each data input $x$. This approach aims to minimize blind drift towards observed classes, and hence, we expect it to enhance overall generalization.

\begin{figure}
    \centering    
    \includegraphics[width=0.6\textwidth]{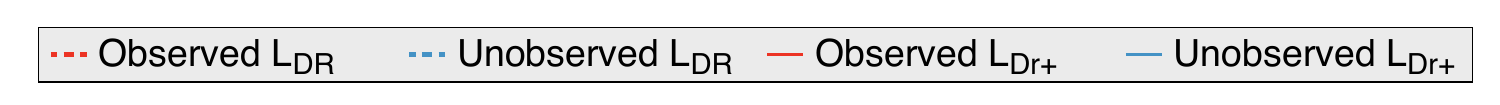} \\
        \begin{subfigure}[b]{0.29\textwidth}
            \centering
            \includegraphics[width=\textwidth]{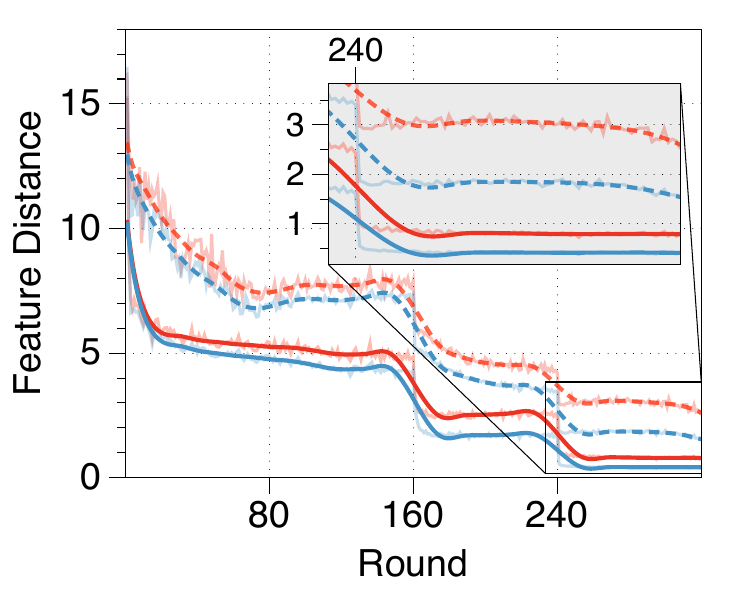}
            \vspace*{-15pt}
            \subcaption{Feature distance}
            \label{fig:feature_distance_dynamic}
        \end{subfigure}
        \hfill
        \begin{subfigure}[b]{0.29\textwidth}
            \centering
                \includegraphics[width=\textwidth]{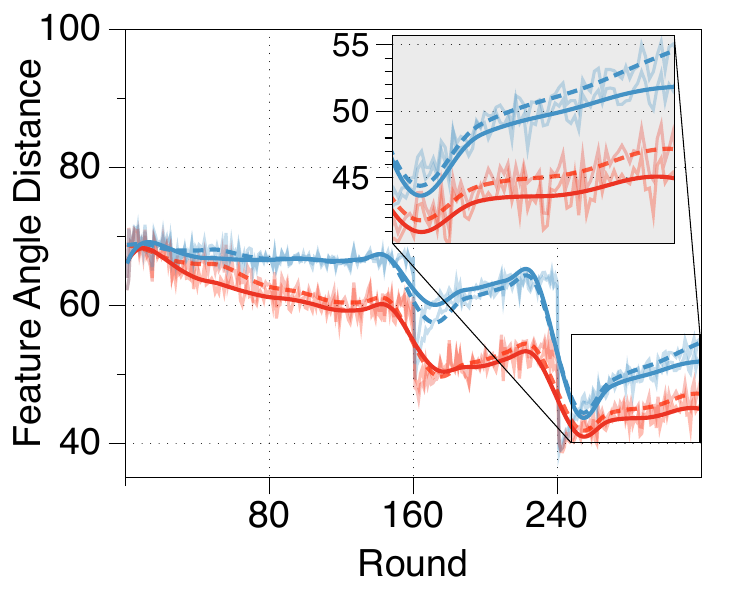}  
            \vspace*{-15pt}
            \subcaption{Feature angle distance}
            \label{fig:feature_angle_dynamics}
        \end{subfigure}
        \hfill
        \begin{subfigure}[b]{0.29\textwidth}
            \centering
                \includegraphics[width=\textwidth]{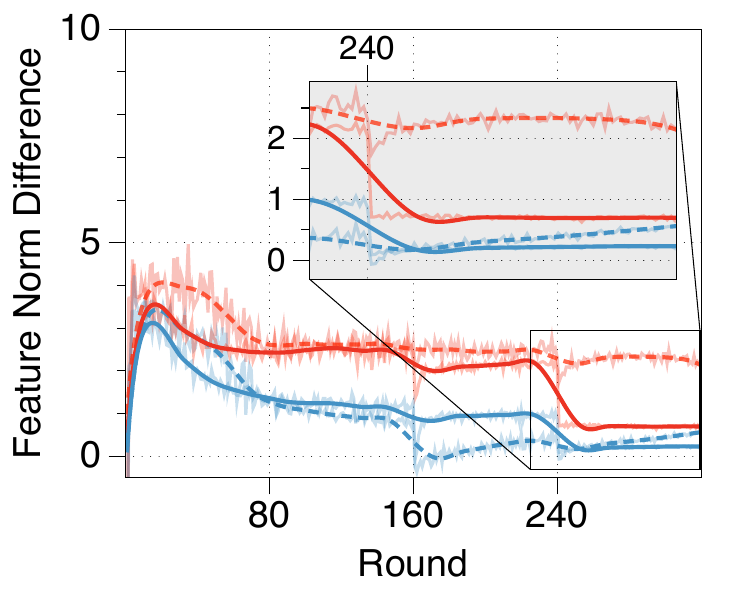} 
            \vspace*{-15pt}
            \subcaption{Feature norm difference}
            \label{fig:feature_norm_dynamics}
        \end{subfigure}
        \caption{We present (a) feature distance, (b) feature angle distance, (c) and feature norm difference from $\bm{\theta}_{r-1}^g$ to $\bm{\theta}_{r}^i$ for \textcolor{red}{observed} and \textcolor{blue}{unobserved} classes  by training with $\mathcal{L}_\text{DR}$ and $\mathcal{L}_\text{Dr+}$.}
        \vspace*{-10pt}
        \label{fig:feature_dynamic_change}
        
\end{figure}

\vspace{-8pt}
\subsection{Effect of Feature Distillation}\label{subsec:effect_fd}
\vspace{-4pt}

Our findings from \autoref{subsec:dotreg} indicate that $\mathcal{L}_\text{DR}$ is unsuitable for the heterogeneous FL environment. This is primarily because there is a notable gap in how features align with the fixed classifier between $\mc{O}^i$ and $\mc{U}^i$. To assess the effect of feature distillation ($\mathcal{L}_{\text{FD}}$), which imposes a constraint on the feature distance $\|f(x; \bm{\theta}_{r}^i) - f(x; \bm{\theta}_{r-1}^g)\|_2$ for $x \in \mc{O}^i$, we measure this distance for both $\mc{O}^i$ and $\mc{U}^i$ from the models trained with $\mathcal{L}_{\text{DR}}$ and $\mathcal{L}_{\text{Dr+}}$. We additionally analyze the angle distance, $\angle(f(x; \bm{\theta}_{r}^i), f(x; \bm{\theta}_{r-1}^g))$, and feature norm difference, $\|f(x; \bm{\theta}_{r}^i)\|_2 - \|f(x; \bm{\theta}_{r-1}^g)\|_2$, as these factors influence the feature distance. These values are averaged over the selected client set $\mathcal{S}_r$.

\myparagraph{Feature distillation stabilizes the feature dynamics.}
By adding $\mathcal{L}_\text{FD}$, as revealed in \autoref{fig:feature_distance_dynamic}, the local model trained with $\mathcal{L}_\text{Dr+}$ shows a reduction in feature distance for \textcolor{red}{observed} classes, compared to the model trained with $\mathcal{L}_\text{DR}$. This reduction happens even for \textcolor{blue}{unobserved} classes. As demonstrated in~\autoref{fig:feature_angle_dynamics} and~\autoref{fig:feature_norm_dynamics}, reduction of feature distance originates from reducing the feature angle distance and feature norm difference for both class sets. In both local models trained with $\mathcal{L}_\text{DR}$ and $\mathcal{L}_\text{Dr+}$, there is a trend where the angle is significantly larger for $\mc{U}^i$ than for $\mc{O}^i$~(\autoref{fig:feature_angle_dynamics}), while the norm difference is smaller for $\mc{U}^i$ than for $\mc{O}^i$~(\autoref{fig:feature_norm_dynamics}). This large angle distance of $\mc{U}^i$ leads to the degradation of the feature-classifier alignment. By minimizing the angle distance via feature distillation, the global model's accuracy improved substantially, rising from $42.52\%$ with $\mc{L}_{\text{DR}}$ to $48.69\%$with $\mc{L}_{\text{Dr+}}$.

\myparagraph{Stabilized features enhance alignment and accuracy.}

We confirm that feature distillation term $\mc{L}_{\text{FD}}$ stabilizes feature dynamics for both $\mc{O}^i$ and $\mc{U}^i$, enhancing the global model's capabilities. While the feature difference is stabilized via $\mc{L}_{\text{FD}}$, it is essential to verify whether this leads to improved alignment and accuracy.
\newpage
\begin{wrapfigure}{r}{0.5\linewidth}
  \centering
  \small
  \vspace*{-5pt}
  \includegraphics[width=0.9\linewidth]{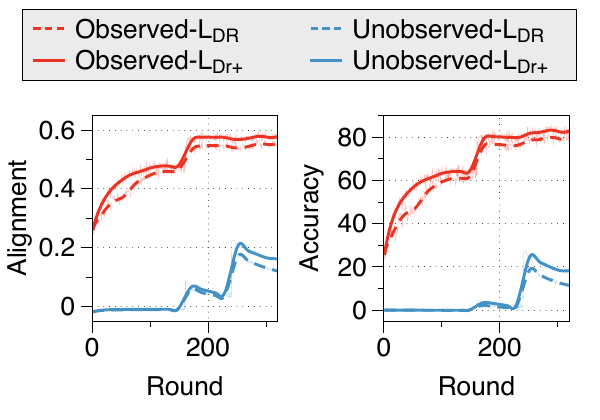}
  \vspace*{-10pt}
  \caption{Comparison of alignment/accuracy on the \textcolor{red}{observed} and \textcolor{blue}{unobserved} classes test data for $\bm{\theta}_r^i$ trained with $\mathcal{L}_\text{DR}$ and $\mathcal{L}_\text{Dr+}$.}
  \vspace{-20pt}
  \label{fig:comparison}
\end{wrapfigure}
In~\autoref{fig:comparison}, we examine in both aspects and illustrate the training curve.  Our proposed algorithm, \ie $\mc{L}_{\text{Dr+}}$, demonstrates superior performance for both $\mc{O}^i$ and $\mc{U}^i$ in terms of alignment and accuracy. Notably, even with the addition of a term to the dot-regression loss, alignment is improved. We attribute this improvement to the enhanced knowledge of the global model which is preserved by preventing the forgetting of previously trained knowledge. Even though the proposed regularizer demonstrates a reasonable regularizing effect, one question remains: ``\emph{Is it superior to other previously used regularizers?}''

\begin{table}[!t]
    \centering
    \vspace{-0.15 in}
    \caption{Synergy of various FL algorithms and regularizers. Baseline indicates training FL models without a regularizer. FD denotes feature distillation, which is the regularizer we use in \alg.}
    \label{tab:synergy_effect}
    \vspace{5pt}
    \small
    \addtolength{\tabcolsep}{-1pt}
    \resizebox{\textwidth}{!}{
    \begin{tabular}{l|cccccc|cccccc}
      \toprule 
       & \multicolumn{6}{c|}{Sharding ($s=10$)} & \multicolumn{6}{c}{LDA ($\alpha=0.1$)} \\ \cmidrule{2-13}
      Algorithm & Baseline & \!\!+Prox\,\cite{li2020federated}\!\! & \!\!+KD\,\cite{hinton2015distilling}\!\! & \!\!+NTD\,\cite{lee2022preservation}\!\! & \!\!+LD\,\cite{kim2021comparing}\!\! & +FD & Baseline & \!\!+Prox\,\cite{li2020federated}\!\! & \!\!+KD\,\cite{hinton2015distilling}\!\! & \!\!+NTD\,\cite{lee2022preservation}\!\! & \!\!+LD\,\cite{kim2021comparing}\!\! & +FD\\ \midrule
      FedAvg~\cite{mcmahan2017communication} & 37.22 &  30.27 & 35.14 &  35.56 & 34.83 &  37.82 & 42.52 &  36.09 & 41.48 & 41.34 & 43.36 & 43.10\\ \midrule      
      FedBABU~\cite{oh2021fedbabu} & 46.20 &  36.71 & 45.50 &  45.09 & 45.81 &  45.31 & 47.37 &  39.04 & 45.58 & 45.56 & 46.46 & 44.77\\ 
   
      SphereFed~\cite{dong2022spherefed} & 43.90 &  1.36 &  41.01 & 43.47 &  41.73 & 45.21 &  46.98 & 1.46 &  45.22 & 46.25 & 43.84 &  48.61\\
      
      FedETF~\cite{li2023no} & 32.42 &  25.18 & 32.76 &  31.98 & 32.25 &  32.77 & 46.27 &  34.92 & 44.94 & 45.77 & 44.36 & 45.92 \\ \midrule
      
      Dot-Regression & 42.52 & 5.42 &  46.60 & 45.78 &  47.52 & \textbf{48.69} &  42.72  &  7.47 & 48.19 & 33.08 & 49.09 & \textbf{50.79} \\ 
      
      \bottomrule
    \end{tabular}
    }
    \vspace{-5pt}
\end{table}

\subsection{Synergistic Effect with Different Types of FL Algorithms and Regularizers}\label{subsec:synergy_effect}
We answer the above question by evaluating the synergy effect of various FL algorithms by maintaining their original training loss and incorporating specific regularizers, as suggested in \autoref{alg:FedDR}. Our study includes FedAvg~\cite{mcmahan2017communication} without classifier freezing and other advanced frameworks such as FedBABU~\cite{oh2021fedbabu}, SphereFed~\cite{dong2022spherefed}, FedETF~\cite{li2023no}, and dot-regression, all of which update local models while freezing the classifier. In addition to the FD regularizer, we consider regularizers such as Prox~\cite{li2020federated} to constrain the distance between local and global model parameters, and several logit-based regularizers---KD~\cite{hinton2015distilling}, NTD~\cite{lee2022preservation, zhao2022decoupled}, and LD~\cite{kim2021comparing}---to keep logit-related measurement of local models from deviating significantly from that of the global model. Specifically, KD applies the softened softmax probability from the logit vector, NTD does the same but excludes the true class dimension, and LD distills the entire logit vector.

\autoref{tab:synergy_effect} demonstrates that \alg\,(dot-regression\,+\,FD) achieves the best performance. Generally, Prox tends to be less effective than logit-based regularizers, which are often outperformed by FD across most algorithms. This is because, as noted in~\autoref{subsec:method}, with the frozen classifier, features are expected to have rich information to mitigate the drift. Prox uniformly regularizes all data instances, whereas logit and feature regularizers adapt to both model parameters and data instances, offering more refined control. Specifically, FD regularizer, with its higher dimensionality, captures the global model's information more precisely than logit-based ones, resulting in better synergy.

\section{Experiments and Results}
\label{sec:exp}

In this section, we present the experimental results of \alg, encompassing both global federated learning (GFL) and personalized federated learning (PFL).  Additionally, we perform various hyper-parameter sensitivity analyses, exploring the impact of varying local epochs and the client sampling ratio on performance, as well as the effect of different $\beta$ values in \alg.

\subsection{Experimental Setup}

\myparagraph{Dataset and models.}
To simulate a realistic FL scenario involving 100 clients, we conduct extensive studies on two widely used datasets: CIFAR-10 and CIFAR-100~\citep{krizhevsky2009cifar}. For CIFAR-10, we employ VGG11~\citep{simonyan2014very}, while for CIFAR-100, MobileNet~\citep{howard2017mobilenets} is used. The training data is distributed among 100 clients using sharding and the LDA (Latent Dirichlet Allocation) partition strategies. 

Following the convention, sharding distributes the data into non-overlapping shards of equal size, each shard encompassing $\frac{|D_\text{train}|}{100\times s}$ and $\frac{|D_\text{test}|}{100\times s}$ samples per class, where $s$ denotes the number of shards per client. On the other hand, LDA involves sampling a probability vector from Dirichlet distribution, $p_c = (p_{c,1}, p_{c,2}, \cdots, p_{c,100}) \sim \text{Dir}(\alpha)$, and allocating a proportion $p_{c,k}$ of instances of class $c \in [C]$ to each client $k \in [100]$. Smaller values of $s$ and $\alpha$ increase the level of data heterogeneity.

\myparagraph{Implementation details.} In each round of communication, a fraction of clients equal to 0.1 is randomly selected to participate in the training process. The total number of communication rounds is 320. The initial learning rate and the number of local epochs for CIFAR-10 and CIFAR-100 are determined through grid searches, with the detailed process and results provided in Appendix~\ref{appsec:exp_setup}. The learning rate $\eta$ is decayed by a factor of 0.1 at the 160th and 240th communication rounds. The number of local epochs is set to 10 for CIFAR-10 and 3 for CIFAR-100 in the main experiments.\footnote{\autoref{tab:gfl_acc_all} is constructed using the code structure from \code{https://github.com/Lee-Gihun/FedNTD}, while the rest of the implementation is based on \code{https://github.com/jhoon-oh/FedBABU}.}
\vspace{-5pt}
\subsection{Global Federated Learning Results}
\label{subsec:gfl_results}

\begin{table}[!t]
    \centering
    \small
    \caption{Accuracy comparison in the GFL setting. 
    The entries are based on results obtained from three different seeds, indicating the mean and standard deviation of the accuracy of the global model, represented as X{\tiny $\pm$Y}. The best performance in each case is highlighted in \textbf{bold}.}
    \label{tab:gfl_acc_all}
    \vspace{5pt}
    \resizebox{.93\textwidth}{!}{
    \begin{tabular}{l|cccc|ccc}
      \toprule 
      \multicolumn{8}{c}{\textbf{NIID Partition Strategy: Sharding}}\\ \midrule
       & \multicolumn{4}{c|}{MobileNet on CIFAR-100} & \multicolumn{3}{c}{VGG on CIFAR-10}\\  \cmidrule{2-8}
      Algorithm & $s$=10 & $s$=20 & $s$=50 & $s$=100  & $s$=2 & $s$=5 & $s$=10\\ \midrule
      FedAvg~\cite{mcmahan2017communication}  & 36.63{\tiny $\pm$ 0.22} & 42.25{\tiny $\pm$ 1.42} & 45.57{\tiny $\pm$ 0.22} & 48.20{\tiny $\pm$ 1.36} & 72.08{\tiny $\pm$ 0.67} & 81.53{\tiny $\pm$ 0.35} & 82.38{\tiny $\pm$ 0.40}   \\ 

      SCAFFOLD~\cite{karimireddy2020scaffold}\,($\times 3$)\!\!
      & 46.08{\tiny $\pm$ 0.37} & 48.15{\tiny $\pm$ 1.21} & 49.31{\tiny $\pm$ 0.62} & 50.73{\tiny $\pm$ 0.42} & 75.49{\tiny $\pm$ 0.42} & \textbf{84.14}{\tiny $\pm$ 0.13} & \textbf{85.11}{\tiny $\pm$ 0.29}   \\ 

      FedNTD~\cite{lee2022preservation}  & 34.05{\tiny $\pm$ 1.19} & 41.78{\tiny $\pm$ 0.31} & 46.42{\tiny $\pm$ 0.63} & 47.17{\tiny $\pm$ 0.32} & 72.21{\tiny $\pm$ 0.59} & 69.96{\tiny $\pm$ 17.10} & 81.99{\tiny $\pm$ 0.42}   \\ 
      FedExP~\cite{jhunjhunwala2023fedexp}  & 36.85{\tiny $\pm$ 0.11} & 42.49{\tiny $\pm$ 1.22} & 45.07{\tiny $\pm$ 0.92} & 48.09{\tiny $\pm$ 1.00} & 72.31{\tiny $\pm$ 0.60} & 81.41{\tiny $\pm$ 0.19} & 82.47{\tiny $\pm$ 0.16}   \\ \midrule

      FedBABU~\cite{oh2021fedbabu}  & 45.97{\tiny $\pm$ 0.48} & 45.53{\tiny $\pm$ 0.79} & 46.52{\tiny $\pm$ 0.51} & 46.02{\tiny $\pm$ 0.28} & 71.99{\tiny $\pm$ 0.52} & 81.07{\tiny $\pm$ 0.60} & 82.32{\tiny $\pm$ 0.06}   \\ 
      
      SphereFed~\cite{dong2022spherefed} & 42.71{\tiny $\pm$ 0.65} & 48.63{\tiny $\pm$ 0.90} & {\textbf{52.16}}{\tiny $\pm$ 0.22} & \textbf{53.41}{\tiny $\pm$ 0.19} & \textbf{76.33}{\tiny $\pm$ 0.33} & 83.67{\tiny $\pm$ 0.18} & 84.36{\tiny $\pm$ 0.30}   \\ 
      FedETF~\cite{li2023no} & 31.37{\tiny $\pm$ 0.72} & 42.22{\tiny $\pm$ 0.77} & 47.47{\tiny $\pm$ 0.67} & 49.00{\tiny $\pm$ 0.74} & 67.81{\tiny $\pm$ 0.94} & 80.78{\tiny $\pm$ 0.68} & 82.60{\tiny $\pm$ 0.46}   \\ 
      \alg \textbf{(Ours)} & \textbf{48.21}{\tiny $\pm$ 0.56} & \textbf{50.77}{\tiny $\pm$ 0.14} & {\textbf{52.15}}{\tiny $\pm$ 0.03} & \textbf{52.41}{\tiny $\pm$ 0.81} & \textbf{76.57}{\tiny $\pm$ 0.51} & 83.22{\tiny $\pm$ 0.34} & 84.14{\tiny $\pm$ 0.27}   \\ \bottomrule
      \midrule
      \multicolumn{8}{c}{\textbf{NIID Partition Strategy: LDA}}\\ \midrule
        & \multicolumn{4}{c|}{MobileNet on CIFAR-100} & \multicolumn{3}{c}{VGG on CIFAR-10} \\  \cmidrule{2-8}
      Algorithm & $\alpha$=0.05 & $\alpha$=0.1 & $\alpha$=0.2 & $\alpha$=0.3  & $\alpha$=0.1 & $\alpha$=0.2 & $\alpha$=0.3\\ \midrule
      FedAvg~\cite{mcmahan2017communication}  & 35.58{\tiny $\pm$ 1.35} & 42.10{\tiny $\pm$ 0.60} & 44.78{\tiny $\pm$ 0.72} & 45.73{\tiny $\pm$ 0.88} & 68.71{\tiny $\pm$ 1.82} & 77.75{\tiny $\pm$ 0.26} & 80.76{\tiny $\pm$ 0.51}   \\ 

      SCAFFOLD~\cite{karimireddy2020scaffold}\,($\times 3$)\!\! & 40.54{\tiny $\pm$ 0.48} & 46.14{\tiny $\pm$ 0.70} & 47.98{\tiny $\pm$ 0.93} & 48.06{\tiny $\pm$ 1.08} & \textit{(Failed)} & 80.15{\tiny $\pm$ 0.29} & {\textbf{82.63}}{\tiny $\pm$ 0.23}   \\ 

      FedNTD~\cite{lee2022preservation}  & 31.78{\tiny $\pm$ 3.14} & 40.41{\tiny $\pm$ 0.96} & 43.10{\tiny $\pm$ 2.03} & 43.04{\tiny $\pm$ 0.82} & 70.22{\tiny $\pm$ 0.40} & 77.16{\tiny $\pm$ 0.20} & 79.50{\tiny $\pm$ 0.56}   \\ 
      FedExP~\cite{jhunjhunwala2023fedexp}  & 34.39{\tiny $\pm$ 1.77} & 40.85{\tiny $\pm$ 1.32} & 44.47{\tiny $\pm$ 0.28} & 45.44{\tiny $\pm$ 0.14} & 70.14{\tiny $\pm$ 0.53} & 78.09{\tiny $\pm$ 0.21} & 80.40{\tiny $\pm$ 0.54}   \\ \midrule

      FedBABU~\cite{oh2021fedbabu}  & 41.97{\tiny $\pm$ 1.01} & 45.77{\tiny $\pm$ 0.28} & 44.28{\tiny $\pm$ 0.45} & 44.80{\tiny $\pm$ 0.63} & 65.15{\tiny $\pm$ 3.66} & 77.03{\tiny $\pm$ 0.25} & 79.91{\tiny $\pm$ 0.13}   \\ 
      
      SphereFed~\cite{dong2022spherefed} & 39.56{\tiny $\pm$ 0.48} & 46.54{\tiny $\pm$ 0.58} & \textbf{49.41}{\tiny $\pm$ 0.78} & 49.22{\tiny $\pm$ 0.86} & 67.49{\tiny $\pm$ 3.49} & 80.05{\tiny $\pm$ 0.40} & {\textbf{82.62}}{\tiny $\pm$ 0.66}   \\ 
      FedETF~\cite{li2023no} & 40.71{\tiny $\pm$ 0.90} & 45.63{\tiny $\pm$ 0.33} & 46.28{\tiny $\pm$ 1.05} & 46.69{\tiny $\pm$ 0.87} & 70.75{\tiny $\pm$ 0.36} & 77.86{\tiny $\pm$ 0.46} & 79.95{\tiny $\pm$ 0.34}   \\ 
      \alg \textbf{(Ours)} & \textbf{45.12}{\tiny $\pm$ 1.00} & \textbf{49.48}{\tiny $\pm$ 0.50} & \textbf{50.67}{\tiny $\pm$ 0.88} & \textbf{51.15}{\tiny $\pm$ 0.65} & \textbf{72.07}{\tiny $\pm$ 2.26} & \textbf{80.90}{\tiny $\pm$ 0.02} & {\textbf{82.42}}{\tiny $\pm$ 0.10}   \\ \midrule
    \end{tabular}
    }
\end{table}

We compare \alg with a range of GFL algorithms, considering both non-freezing and freezing classifier approaches. Among non-freezing classifiers, \alg competes with FedAvg~\cite{mcmahan2017communication}, SCAFFOLD~\cite{karimireddy2020scaffold}, FedNTD~\cite{lee2022preservation}, and FedExP~\cite{jhunjhunwala2023fedexp}. \alg is also evaluated against freezing classifier algorithms such as FedBABU~\cite{oh2021fedbabu}, SphereFed~\cite{dong2022spherefed}, and FedETF~\cite{li2023no}. Among the baseline algorithms, SCAFFOLD incurs a communication cost three times higher per round, denoted as ($\times3$).  Our experiments encompass heterogeneous settings involving sharding and LDA non-IID environments. 

\autoref{tab:gfl_acc_all} summarizes the accuracy comparison between the state-of-the-art GFL methods and FedAvg under various conditions.
While specific methods demonstrated effectiveness in particular scenarios, some of these frequently underperformed relative to the robustness of FedAvg. For example, SCAFFOLD shown strong performance in the less heterogeneous sharding setting on CIFAR-10; however, it failed in model training under the highly heterogeneous LDA condition with $\alpha=0.1$.
 Notably, \alg consistently exceeded all baseline methods in performance across diverse experimental conditions and often achieved state-of-the-art results. \alg demonstrated exceptional performance in highly heterogeneous FL environments, particularly excelling in the CIFAR-100 LDA configuration with $\alpha=0.05$, achieving a notable 3.15\% improvement over all baseline models.

\subsection{Personalized Federated Learning Results}
\label{subsec:pfl_results}

\begin{table}[t!]
    \centering
    \caption{PFL accuracy comparison with MobileNet on CIFAR-100. For PFL, we denote the entries in the form of X{\tiny $\pm$(Y)}, representing the mean and standard deviation of personalized accuracies across all clients derived from a single seed.}
    \label{tab:pfl_acc}
    \small
    \vspace{5pt}
    \resizebox{0.95\textwidth}{!}{
    \begin{tabular}{l|ccc|ccc}
    \toprule
    Algorithm & $s$=10 & $s$=20 & $s$=100 & $\alpha$=0.05 & $\alpha$=0.1 & $\alpha$=0.3 \\ \midrule
    Local only ($\mathcal{L}_\text{CE}$)                              & 58.05{\tiny $\pm$(8.11)}  & 42.45{\tiny $\pm$(6.44)} & 18.69{\tiny $\pm$(3.28)} & 55.39{\tiny $\pm$(8.79)}   & 43.76{\tiny $\pm$(7.46)} & 27.75{\tiny $\pm$(5.32)} \\
    
    Local only ($\mathcal{L}_\text{CE}$+ETF)
     & 58.01{\tiny $\pm$(7.34)}   & 41.62{\tiny $\pm$(5.91)} & 18.92{\tiny $\pm$(3.00)} & 55.34{\tiny $\pm$(9.13)}  & 43.37{\tiny $\pm$(7.12)} & 27.87{\tiny $\pm$(5.34)} \\ 
        

    Local only ($\mathcal{L}_\text{DR}$)                               & 60.68{\tiny $\pm$(7.77)}   & 44.61{\tiny $\pm$(6.61)} & 20.98{\tiny $\pm$(3.49)} & 58.56{\tiny $\pm$(9.16)}  & 46.72{\tiny $\pm$(7.29)} & 30.88{\tiny $\pm$(5.33)} \\    
    
    \midrule
    FedPer~\cite{arivazhagan2019federated}                            & 70.67{\tiny $\pm$(7.19)}  & 57.27{\tiny $\pm$6.66} & 24.30{\tiny $\pm$(4.34)} & 62.67{\tiny $\pm$(7.65)}  & 53.43{\tiny $\pm$(6.60)} & 35.68{\tiny $\pm$(4.82)} \\ 
    Per-FedAvg~\cite{fallah2020personalized}                            & 32.13{\tiny $\pm$(10.90)}  & 36.66{\tiny $\pm$(8.86)} & 41.27{\tiny $\pm$(7.43)} & 28.81{\tiny $\pm$(8.68)}  & 35.56{\tiny $\pm$(6.56)} & 42.80{\tiny $\pm$(4.76)} \\ 
    
    FedRep~\cite{collins2021exploiting}                            & 63.14{\tiny $\pm$(7.63)}  & 51.69{\tiny $\pm$(6.50)} & 26.31{\tiny $\pm$(4.74)} &  57.53{\tiny $\pm$(8.05)}  & 49.60{\tiny $\pm$(6.25)} & 37.00{\tiny $\pm$(4.82)} \\ 
    
    Ditto~\cite{li2021ditto}                            & 39.26{\tiny $\pm$(14.43)}  & 38.18{\tiny $\pm$(9.96)} & 44.53{\tiny $\pm$(5.08)} & 35.81{\tiny $\pm$(14.83)}  & 37.81{\tiny $\pm$(11.80)} & 43.72{\tiny $\pm$(5.12)} \\ \midrule

    FedAvg-FT~\cite{mcmahan2017communication} & 69.81{\tiny $\pm$(6.78)}  & 56.13{\tiny $\pm$(5.77)} & 47.66{\tiny $\pm$(5.20)} & 63.37{\tiny $\pm$(9.28)}  & 56.79{\tiny $\pm$(5.96)} & 50.12{\tiny $\pm$(3.67)} \\ 

    FedBABU-FT~\cite{oh2021fedbabu} & 80.14{\tiny $\pm$(6.25)}  & 70.89{\tiny $\pm$(5.60)} & 52.14{\tiny $\pm$(5.09)} & 75.50{\tiny $\pm$(6.40)}  & 70.83{\tiny $\pm$(5.06)} & 56.91{\tiny $\pm$(3.74)} \\
    
    SphereFed-FT~\cite{dong2022spherefed} & 81.90{\tiny $\pm$(5.86)}  & 71.56{\tiny $\pm$(5.78)} & 55.83{\tiny $\pm$(4.67)} & 73.21{\tiny $\pm$(7.08)}  & 70.00{\tiny $\pm$(5.09)} & 60.03{\tiny $\pm$(3.99)} \\ 
    
    FedETF-FT~\cite{li2023no} & 53.75{\tiny $\pm$(7.35)}  & 52.94{\tiny $\pm$(5.71)} & 51.69{\tiny $\pm$(5.03)} & 52.96{\tiny $\pm$(8.01)}  & 53.97{\tiny $\pm$(5.40)} & 51.67{\tiny $\pm$(3.83)} \\ \midrule

    \textbf{{\alg}\,FT (ours)} & \textbf{84.10}{\tiny $\pm$(5.43)}  & \textbf{75.42}{\tiny $\pm$(4.80)} & \textbf{56.76}{\tiny $\pm$(4.91)} & \textbf{78.55}{\tiny $\pm$(6.16)}  & \textbf{74.75}{\tiny $\pm$(4.75)} & \textbf{62.16}{\tiny $\pm$(3.73)} \\ 
    \bottomrule
    \end{tabular}
    }
\end{table}

We introduce {\alg}\,FT, inspired by prior work~\citep{oh2021fedbabu, dong2022spherefed, li2023no, kim2023fedfn}, which enhances personalization by leveraging local data to fine-tune the global federated learning (GFL) model. We fine-tune the \alg GFL model using $\mathcal{L}_\text{Dr+}$ to create {\alg}\,FT, \ie 2-step approach. For a comprehensive analysis, we compare {\alg}\,FT with existing personalized federated learning (PFL) methods, including 1-step approaches, \ie creating PFL models from scratch, such as FedPer~\cite{arivazhagan2019federated}, Per-FedAvg~\cite{fallah2020personalized}, FedRep~\cite{collins2021exploiting}, and Ditto~\cite{li2021ditto}, as well as 2-step methods such as FedAVG-FT, FedBABU-FT~\cite{oh2021fedbabu}, SphereFed-FT~\cite{dong2022spherefed}, and FedETF-FT~\cite{li2023no}. Additionally, we compare these methods with various simple local models that have not undergone federated learning: (1) Local only ($\mathcal{L}_\text{CE}$), trained with $\mathcal{L}_\text{CE}$, (2) Local only ($\mathcal{L}_\text{CE}$\,+\,ETF), trained with $\mathcal{L}_\text{CE}$ and initializing the classifier with an ETF classifier, and (3) Local only ($\mathcal{L}_\text{DR}$), trained using $\mathcal{L}_\text{DR}$. 

In \autoref{tab:pfl_acc}, we first compare the performance of simple local models in PFL by examining $\mathcal{L}_\text{DR}$ and $\mathcal{L}_\text{CE}$. While methods using $\mathcal{L}_\text{CE}$ show no significant differences, utilizing $\mathcal{L}_\text{DR}$ leads to substantial performance improvements in PFL across all settings. The ``Local only ($\mathcal{L}_\text{CE}$)'' and ``Local only ($\mathcal{L}_\text{CE}$\,+\,ETF)'' methods exhibit similar performance due to the nearly classwise orthogonal nature of randomly initialized classifiers~\cite{oh2021fedbabu, saxe2013exact, glorot2010understanding, he2015delving, lezama2018ole}. With a large number of classes ($C$=100), the ETF classifier, which is also nearly classwise orthogonal, performs similarly to random initialization. When comparing {\alg}\,FT with other 2-step methods, {\alg}\,FT consistently demonstrates superior performance. This aligns with previous research~\cite{nguyen2022begin,chen2022importance} suggesting that fine-tuning from a well-initialized model yields better PFL performance. Additionally, compared with 1-step algorithms, {\alg}\,FT continues to show superiority, outperforming all baseline methods across all settings.

\subsection{Sensitivity Analysis}\label{subsec:sensitivity}

\begin{figure}\label{fig: Sensitivity}
    \centering
    \vspace{5pt}
    \includegraphics[width=0.4\textwidth]{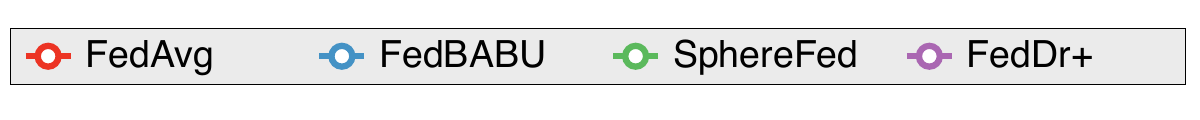} \\
    \begin{subfigure}[b]{0.3\textwidth}
        \centering
        \includegraphics[width=1\textwidth]{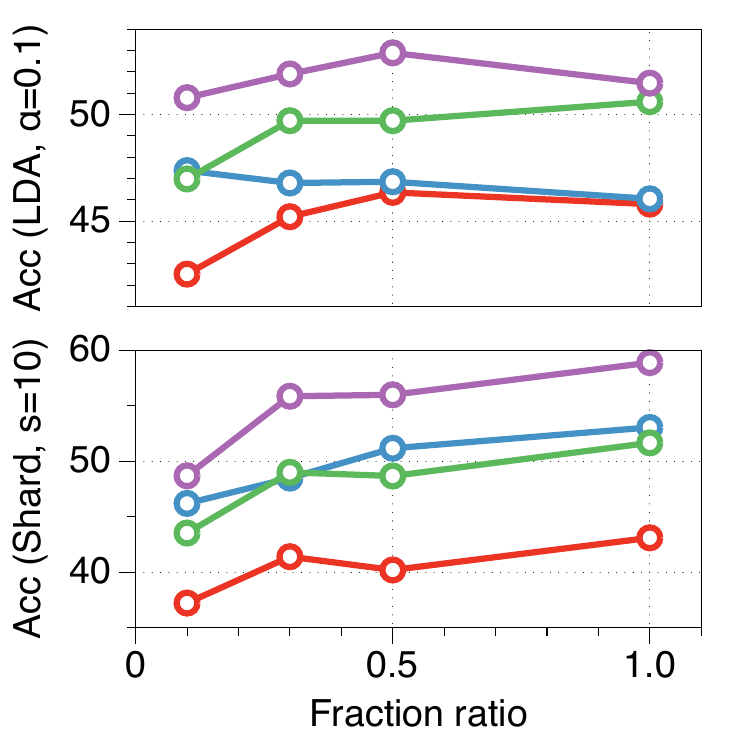}
        \vspace{-15pt}
        \subcaption{Client sampling ratio}
    \end{subfigure}
    \hfill
    \begin{subfigure}[b]{0.3\textwidth}
        \centering
        \includegraphics[width=1\textwidth]{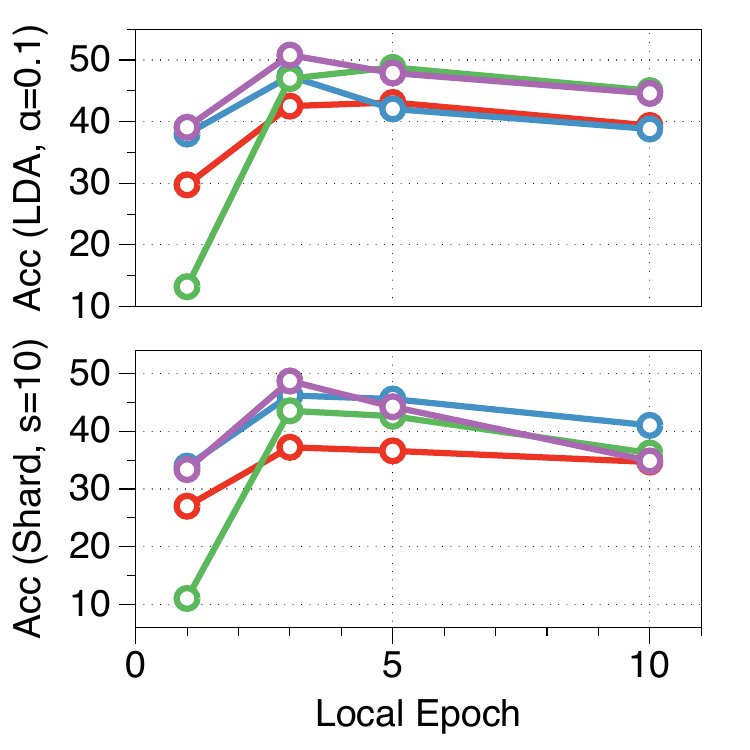}
        \vspace{-15pt}
        \subcaption{Local epochs}
    \end{subfigure}
    \hfill
    \begin{subfigure}[b]{0.3\textwidth}
        \centering
        \includegraphics[width=1\textwidth]{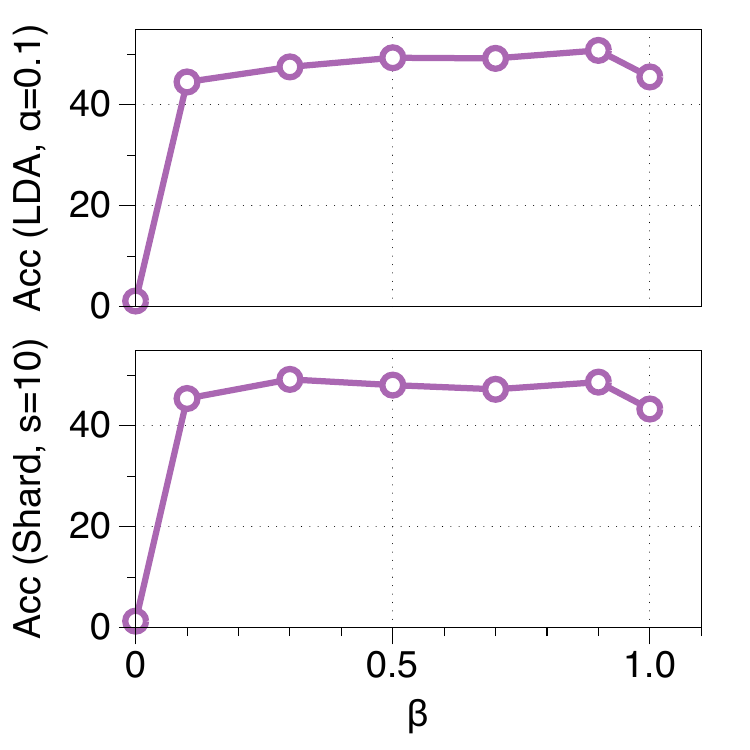}
        \vspace{-15pt}
        \subcaption{$\beta$ sensitivity}
    \end{subfigure}
    \caption{Performance of baselines and \alg on CIFAR-100 ($\alpha$=0.1 and $s$=10) with various analyses: (a) client sampling ratio, (b) the number of local epochs, and (c) sensitivity to $\beta$.}
    \vspace*{-5pt}
    \label{fig:sensitivity}
\end{figure}

We explore the impact of varying client sampling ratio and local epochs on performance, as well as the effect of different $\beta$ values in \alg, as detailed in \autoref{fig:sensitivity}. All experiments are conducted on MobileNet using the CIFAR-100 dataset with sharding ($s$=10) and LDA ($\alpha$=0.1).

\myparagraph{Effect of client sampling ratio and local epochs.} We evaluate the sensitivity of hyperparameters in \alg by comparing it to baselines under varying client sampling ratio and local epochs, starting from the default setting of client sampling ratio of 0.1 and local epoch of 3. Compared to FedAvg (without classifier freezing), FedBABU and SphereFed (all with classifier freezing) show performance improvements with increasing fraction ratios, but \alg consistently outperforms the baselines. The number of local epochs is crucial in FL; too few epochs result in underfitting, while too many cause client drift, degrading global model performance. The default setting of local epochs 3 is optimal for all baselines, with \alg achieving the best performance. Although performance generally declines when deviating from this peak point, \alg remains the best or highly competitive.

\myparagraph{Weight ratio $\beta$ analysis.} We analyze the effect of scaling parameter in \alg by varying $\beta$ while keeping other hyperparameters constant. The performance is evaluated for $\beta\in\{0, 0.1, 0.3, 0.5, 0.7, 0.9, 1.0\}$. When $\beta=0$, only feature distillation is applied, and when $\beta=1$, only dot-regression is used. $\beta\in\{0, 1\}$ are generally less effective, whereas $\beta\in\{0.3, 0.5, 0.7, 0.9\}$ show consistently good performance, indicating a balanced approach is beneficial.

\vspace{-3pt}
\section{Related Work}
\label{sec:related}
\vspace{-5pt}


\myparagraph{Federated learning.}
Federated Learning (FL) is a decentralized approach to deep learning where multiple clients collaboratively train a global model using their own datasets~\cite{mcmahan2017communication, oh2021fedbabu, li2020federated, acar2021federated}. This approach faces challenges due to data heterogeneity across clients, causing instability in the learning process~\cite{karimireddy2020scaffold, luo2021no}. To address this problem, strategies like classifier variance reduction in FedPVR~\cite{li2022effectiveness} and virtual features in CCVR~\cite{luo2021no} have been proposed. Additionally, it is essential to distinguish between Global Federated Learning (GFL) and Personalized Federated Learning (PFL), as these are crucial concepts in FL. GFL aims to improve a single global model's performance across clients by addressing data heterogeneity through methods like client drift mitigation~\cite{li2020federated, karimireddy2020scaffold, jhunjhunwala2023fedexp}, enhanced aggregation schemes~\cite{wang2020federated, wang2020tackling}, and data sharing techniques using public or synthesized datasets~\cite{lin2020ensemble, luo2021no, zhao2018federated}. Otherwise, PFL focuses on creating personalized models for individual clients by decoupling feature extractors and classifiers for unique updates~\cite{oh2021fedbabu, arivazhagan2019federated, collins2021exploiting}, modifying local loss functions~\cite{fallah2020personalized, li2021ditto}, and using prototype communication techniques~\cite{tan2022fedproto, xu2023personalized}.


\myparagraph{Frozen classifier in FL.}
By focusing on alignment, previous studies have attempted to mitigate data heterogeneity by freezing the classifier \cite{oh2021fedbabu, dong2022spherefed, li2023no}. Nevertheless, these methods have yet to effectively improve the alignment between features and their corresponding classifier weights. Motivated by this, we integrated the dot-regression method into federated learning to achieve a better-aligned local model by freezing the classifier. Dot-regression, proposed to address class imbalance, focuses on aligning feature vectors to a fixed classifier, demonstrating superior alignment performance compared to previous approaches. However, optimizing the dot-regression loss to align feature vectors with a fixed classifier caused the local model to lose information on unobserved classes, thereby degrading global model performance. To address these issues, FedLoGe~\cite{xiao2024fedloge} employing realignment techniques to ensure the well-aligned local model's performance translated to the global model. Additionally, in FedGELA~\cite{fan2024federated}, the classifier is globally fixed as a simplex ETF while being locally adapted to personal distributions. Also, FedPAC~\cite{xu2023personalized} addressed these challenges by leveraging global semantic knowledge for explicit local-global feature alignment. Besides alignment-focused methods, there have been various attempts to maintain good local model performance in the global model~\cite{jiang2023heterogeneous, an2024federated, chen2021bridging}. 

\myparagraph{Knowledge distillation in FL.}
Multiple studies have explored knowledge transfer techniques ~\cite{itahara2021distillation, ye2023fake, cho2022heterogeneous, wang2024dfrd, cai2024fed}. Specifically, knowledge distillation (KD) has been widely studied in FL settings, such as in FedMD~\cite{li2019fedmd} and FedDF~\cite{lin2020ensemble}, where a pretrained teacher model transfers knowledge to a student model. Additional distillation-based methods, such as FedFed~\cite{yang2024fedfed} and co-distillation framework for PFL~\cite{chen2024spectral, cho2023communication}, have also been explored. In contrast to existing methods, we propose a loss function incorporating feature distillation to maintain the performance of both local and global models. To our knowledge, this is the first application of feature distillation in FL. This approach highlights the importance of distinguishing between GFL and PFL.

\vspace{-0.1 in}
\section{Conclusion}
\label{sec:conclusion}
\vspace{-0.1 in}




Motivated by the recent FL methods enhancing feature alignment with a fixed classifier, we first investigate the effects of applying dot-regression loss for FL. Since the dot-regression is the most direct method for feature-classifier alignment, we find it improves alignment and accuracy in local models but degrades the performance of the global model. This happens because local clients trained with dot-regression tend to forget classes that have not been observed. To address this, we propose \alg, combining dot-regression with 
a feature distillation method. By regularizing the deviation of local features from global features, \alg allows local models to maintain knowledge about all classes during training, thereby ultimately preserving general knowledge of the global model. Our method achieves top performance in global and personalized FL experiments, even when data is distributed unevenly across devices (non-IID settings).

\bibliographystyle{unsrt} 
\bibliography{neurips_2024}


\newpage
\appendix
\appendix
\supptitle



We organized notations at \autoref{app:notations}. In \autoref{app:prelim}, we show the pulling and pushing gradients of the CE loss in detail.
Then, we elucidate the experimental setup in \autoref{appsec:exp_setup}, encompassing dataset description, model specifications, NIID partition, and hyperparameter search. In \autoref{appsec:add_exp}, we present additional experiment results of PFL and elapsed time measurement. 

\section{Notations}
\label{app:notations}

\begin{table}[htp]
\centering
\vspace{-10pt}
\caption{Notations used throughout the paper.}
\label{tab:main_notation_summary}
\vspace{5pt}
\resizebox{0.85\textwidth}{!}{
\begin{tabular}{ll}
\toprule
\textbf{Indices} & \\ 
$c \in [C]$  & Index for a class \\
$r \in [R]$ & Index for FL round \\
$i \in [N]$ & Index for a client \\ 
\midrule

\textbf{Dataset} & \\
$D_{\text{train}}^{i}$ & Training dataset for client $i$ \\
$D_{\text{test}}^{i}$ & Test dataset for client $i$ \\
$(x, y) \in D_{\text{train,test}}^{i}\,; (x, y) \sim \mc{D}^{i}$ & Data on client $i$ sampled from distribution $\mc{D}^{i}$ \\
 & ($x$: input data, $y$: class label) \\
$\mc{O}^i$ & Dataset consists of observed classes in client $i$ \\
$\mc{U}^i$ & Dataset consists of unobserved classes in client $i$ \\

\midrule
\textbf{Parameters} & \\ 
$\bm{\theta}$ & Feature extractor weight parameters \\
$\bm{V} = [v_1, \ldots, v_C] \in \mathbb{R}^{C \times d}$ & Classifier weight parameters (frozen during training) \\
$v_c, c \in [C]$ & $c$-th row vector of $\bm{V}$ \\
$\bm{\Theta} = (\bm{\theta}, \bm{V})$ & All model parameters \\
$\bm{\Theta}_{r}^{g} = (\bm{\theta}_{r}^{g}, \bm{V})$ & Aggregated global model parameters at round $r$ \\
$\bm{\Theta}_{r}^{i}= (\bm{\theta}_{r}^{i}, \bm{V})$ & Trained model parameters on client $i$ at round $r$ \\

\midrule
\textbf{Model Forward} & \\
$p(x; \bm{\theta}) \in \mathbb{R}^{C}$ & Softmax probability of input $x$ \\ 
$p_c(x; \bm{\theta}), c \in [C]$ & $c$-th element of $p(x; \bm{\theta})$ \\ 
$\mathcal{L}_{\text{CE}}(x; \theta) = -\log p_{y}(x; \bm{\theta})$ & Cross-entropy loss of input $x$ \\
$f(x; \bm{\theta}) \in \mathbb{R}^{d}$ & Feature vector of input $x$ \\ 
$z(x; \bm{\theta}) = f(x; \bm{\theta}) \bm{V}^\top \in \mathbb{R}^{C}$ & Logit vector of input $x$ \\ 
$z_c(x; \bm{\theta}), c \in [C]$  & $c$-th element of $z(x; \bm{\theta})$ \\
\bottomrule
\end{tabular}
}
\end{table}
\section{Preliminaries: Pulling and Pushing Feature Gradients in CE}
\label{app:prelim}

In this section, we first calculate the classifier gradient for features and introduce the pulling and pushing effects of the cross-entropy objective.

\subsection{Feature Gradient of \texorpdfstring{$\mathcal{L}_{\text{CE}}$}{L\_CE}}\label{appsubsec:prop}


We first provide two lemmas supporting Proposition~\ref{appprop:pull_push}, explaining the behavior of pulling and pushing feature gradients in the cross-entropy (CE) loss.

\begin{lem}\label{applem:derivative_logit}    
    For all $c,c'\in[C], \frac{\partial p_{c'}(x;\bm{\theta})}{\partial z_c(x;\bm{\theta})}=
    \begin{cases}
        p_c(x;\bm{\theta}) \cdot (1-p_c(x;\bm{\theta})) &\text{if }c=c'\\
        -p_c(x;\bm{\theta}) \cdot p_{c'}(x;\bm{\theta}) &\text{else }   
    \end{cases}.$
\end{lem}

\begin{proof}
    Note that $p(x;\bm{\theta})=\bigg[\frac{\exp(z_j(x;\bm{\theta}))}{\sum_{i=1}^{C}\exp(z_i(x;\bm{\theta}))}\bigg]_{j=1}^{C}\in\mathbb{R}^{C}$.
    Then,

    \begin{enumerate}[label=(\roman*)]
        \item $c=c'$ case:
            \begin{align*}
                \frac{\partial p_c(x;\bm{\theta})}{\partial z_c(x;\bm{\theta})}&=\frac{\partial }{\partial z_c(x;\bm{\theta})}\left\{\frac{\exp(z_c(x;\bm{\theta}))}{\sum_{i=1}^{C}\exp(z_i(x;\bm{\theta}))}\right\} \\
                &=\frac{\exp(z_c(x;\bm{\theta}))\left(\sum_{i=1}^{C}\exp(z_i(x;\bm{\theta}))\right)-{\exp(z_c(x;\bm{\theta}))}^2}{\left(\sum_{i=1}^{C}\exp(z_i(x;\bm{\theta}))\right)^2}\\
                              &=p_c(x;\bm{\theta})-{p_c(x;\bm{\theta})}^2 =p_c(x;\bm{\theta})(1-p_c(x;\bm{\theta})).
                \end{align*}
        \item $c\neq c'$ case:
            \begin{align*}
                \frac{\partial p_{c'}(x;\bm{\theta})}{\partial z_c(x;\bm{\theta})}&=\frac{\partial }{\partial z_c(x;\bm{\theta})}\left\{\frac{\exp(z_{c'}(x;\bm{\theta}))}{\sum_{i=1}^{C}\exp(z_i(x;\bm{\theta}))}\right\}=\frac{-\exp(z_{c}(x;\bm{\theta}))\exp(z_{c'}(x;\bm{\theta}))}{\left(\sum_{i=1}^{C}\exp(z_i(x;\bm{\theta}))\right)^2}\\
                              &=-p_c(x;\bm{\theta})p_{c'}(x;\bm{\theta}).
            \end{align*}   
    \end{enumerate} 
\end{proof}

\begin{lem}\label{applem:gradient_logit}
 $\nabla_{z(x;\bm{\theta})}\mathcal{L}_\text{CE}(x,y;\bm{\theta})=p(x;\bm{\theta})-\mathbf{e}_{y}$, where $\mathbf{e}_{y}\in\mathbb{R}^{C}$ is the unit vector with its $y$-th element as 1. 
\end{lem}

\begin{proof}
    \begin{align*}
        \frac{\partial\mathcal{L}_\text{CE}(x,y;\bm{\theta})}{\partial z_c(x;\bm{\theta})}&=- \frac{\partial}{\partial z_c(x;\bm{\theta})}\log p_y(x;\bm{\theta})=-\frac{1}{p_y(x;\bm{\theta})}\frac{\partial p_y(x;\bm{\theta})}{\partial z_c(x;\bm{\theta})}\notag\\
        &\stackrel{(\bigstar)}{=}\begin{cases}
            p_c(x;\bm{\theta})-1 &\text{if }c=y\\
            p_c(x;\bm{\theta}) &\text{else }   
            \end{cases} =p_c(x;\bm{\theta})-\mathds{1}\{c=y\}.
    \end{align*}
    Note that $(\bigstar)$ holds by the Lemma~\ref{applem:derivative_logit}.
    Therefore, the desired result is satisfied.
\end{proof}

\begin{prop}\label{appprop:pull_push}
    Given $(x,y)$, the gradient of the $\mathcal{L}_\text{CE}$ with respect to $f(x;\bm{\theta})$ is given by:
    \begin{equation}
    \nabla_{f(x;\bm{\theta})}\mathcal{L}_\text{CE}(x,y;\bm{\theta})= - (1-p_{y}(x;\bm{\theta}))v_{y} + \sum_{c\in [C]\setminus \{y\}} p_{c}(x;\bm{\theta})v_{c} .
    \end{equation}
\end{prop}

\begin{proof}
\begin{align*}   
        &\nabla_{f(x;\bm{\theta})}\mathcal{L}_\text{CE}(x,y;\bm{\theta})\\
            &\stackrel{(\clubsuit)}{=}\left[\nabla_{f(x;\bm{\theta})}z_{1}(x;\bm{\theta})|\cdots|\nabla_{f(x;\bm{\theta})}z_{C}(x;\bm{\theta})\right]\nabla_{z(x;\bm{\theta})}\mathcal{L}_\text{CE}(x,y;\bm{\theta})\\
            &=\sum_{c=1}^{C}\frac{\partial \mathcal{L}_\text{CE}(x,y;\bm{\theta})}{\partial z_{c}(x;\bm{\theta})}\nabla_{f(x;\bm{\theta})}z_{c}(x;\bm{\theta})\\
            &=\frac{\partial \mathcal{L}_\text{CE}(x,y;\bm{\theta})}{\partial z_{y}(x;\bm{\theta})}\nabla_{f(x;\bm{\theta})}z_{y}(x;\bm{\theta})+\sum_{c\in [C]\setminus \{y\}}\frac{\partial \mathcal{L}_\text{CE}(x,y;\bm{\theta})}{\partial z_{c}(x;\bm{\theta})}\nabla_{f(x;\bm{\theta})}z_{c}(x;\bm{\theta})\\
            &=\frac{\partial \mathcal{L}_\text{CE}(x,y;\bm{\theta})}{\partial z_{y}(x;\bm{\theta})}v_{y}+\sum_{c\in [C]\setminus \{y\}}\frac{\partial \mathcal{L}_\text{CE}(x,y;\bm{\theta})}{\partial z_{c}(x;\bm{\theta})}v_{c}\\
            &\stackrel{(\spadesuit)}{=} - (1-p_{y}(x;\bm{\theta}))v_{y} + \sum_{c\in [C]\setminus \{y\}} p_{c}(x;\bm{\theta})v_{c}.\\   
\end{align*}
Employing the chain rule for $(\clubsuit)$ and invoking Lemma~\ref{applem:gradient_logit} for $(\spadesuit)$ confirms the result.

\end{proof}

\subsection{Physical Meaning of \texorpdfstring{$\nabla_{f(x;{\theta})}\mathcal{L}_\text{CE}(x,y;{\theta})$}{grad\_L\_CE}}

Note that $\nabla_{f(x;\bm{\theta})}\mathcal{L}_\text{CE}(x,y;\bm{\theta})$ has two components: $\mathbf{F}_\text{Pull} = (1 - p_{y}(x;\bm{\theta}))v_{y}$ and $\mathbf{F}_\text{Push} = -\sum_{c \in [C] \setminus \{y\}} p_{c}(x;\bm{\theta})v_{c}$. 
$\mathbf{F}_\text{Pull}$ adjusts the feature vector in the positive direction of the actual class index's classifier vector $v_y$, guiding alignment towards $v_y$. Conversely, $\mathbf{F}_\text{Push}$ adjusts the feature vector in the negative direction of the vectors in the not-true class set $[C] \setminus \{y\}$, inducing misalignment towards $v_c$ for $c \in [C] \setminus \{y\}$.

    


\section{Experimental Setup}\label{appsec:exp_setup}


\subsection{Code Implementation}

Our implementations are conducted using the PyTorch framework. Specifically, the experiments presented in \autoref{tab:gfl_acc_all} are executed on a single NVIDIA RTX 3090 GPU, based on the code structure from the following repository: \url{https://github.com/Lee-Gihun/FedNTD}. The other parts of our study are carried out on a single NVIDIA A5000 GPU, utilizing the code framework from \url{https://github.com/jhoon-oh/FedBABU}.

\subsection{Datasets, Model, and Optimizer}

To simulate a realistic FL scenario, we conduct extensive studies on two widely used datasets: CIFAR-10 and CIFAR-100~\citep{krizhevsky2009cifar}. A momentum optimizer is utilized for all experiments. Unless otherwise noted, the basic setting of our experiments follows the dataset statistics, FL scenario specifications, and optimizer hyperparameters summarized in Table~\ref{apptab:dataset_optimizer_details}.

\begin{table}[h!]
    \centering
    \small
    \caption{Summary of Dataset, Model, FL System, and Optimizer Specifications}
    \label{apptab:dataset_optimizer_details}
    \vspace{5pt}
    \begin{tabular}{l|cccccccccc}
        \toprule
        {Datasets} & $C$ & $|D_\text{train}|$  & $|D_\text{test}|$ & $N$ & $R$ & $r$ & $E$ & $B$ & $m$ & $\lambda$ \\
        \midrule
        CIFAR-10 & 10 & 50000 & 10000 & 100 & 320 & 0.1 & 10 & 50 & 0.9 & 1e-5 \\
        CIFAR-100 & 100 & 50000 & 10000 & 100 & 320 & 0.1 & 3 & 50 & 0.9 & 1e-5 \\

        \bottomrule
    \end{tabular}
\end{table}

Note: In terms of dataset information, $C$ represents the number of classes in the dataset, with $|D_{\text{train}}|$ and $|D_{\text{test}}|$ indicating the total numbers of training and test data used, respectively. For the federated learning (FL) system specifics, $R$ indicates the total number of FL rounds, $r$ is the ratio of clients selected for each round, and $E$ denotes the number of local epochs. Local model training utilizes a momentum optimizer where $B$ is the batch size, and $m$ and $\lambda$ represent the momentum and weight decay parameters, respectively. The initial learning rate $\eta$ is decayed by a factor of 0.1 at the 160th and 240th communication rounds. The initial learning rate $\eta$ and batch size $B$ were determined via extensive grid search for each algorithm, details outlined in Appendix~\ref{appsubsec:grid_search}.

\subsection{Non-IID Partition Strategies}\label{appsubsec:niid_strategy}

To induce heterogeneity in each client's training and test data ($D_\text{train}^i, D_\text{test}^i$), we distribute the entire class-balanced datasets, $D_\text{train}$ and $D_\text{test}$, among 100 clients using both sharding and Latent Dirichlet Allocation (LDA) partitioning strategies:

\begin{itemize}[leftmargin=10pt]
\item \textbf{Sharding}~\cite{mcmahan2017communication, oh2021fedbabu}: We organize the $D_\text{train}$ and $D_\text{test}$ by label and divide them into non-overlapping shards of equal size. Each shard encompasses $\frac{|D_\text{train}|}{100\times s}$ and $\frac{|D_\text{test}|}{100\times s}$ samples of the same class, where $s$ denotes the number of shards per client. This sharding technique is used to create $D_\text{train}^i$ and $D_\text{test}^i$, which are then distributed to each client $i$, ensuring that each client has the same number of training and test samples. The data for each client is disjoint. As a result, each client has access to a maximum of $s$ different classes. Decreasing the number of shards per user $s$ increases the level of data heterogeneity among clients.

\item \textbf{Latent Dirichlet Allocation (LDA)}~\cite{luo2021no, wang2020federated}: We utilize the LDA technique to create $D_\text{train}^i$ from $D_\text{train}$. This involves sampling a probability vector $p_c = (p_{c,1}, p_{c,2}, \cdots, p_{c,100}) \sim \text{Dir}(\alpha)$ and allocating a proportion $p_{c,k}$ of instances of class $c \in [C]$ to each client $k \in [100]$. Here, $Dir(\alpha)$ represents the Dirichlet distribution with the concentration parameter $\alpha$. The parameter $\alpha$ controls the strength of data heterogeneity, with smaller values leading to stronger heterogeneity among clients. For $D_\text{test}^i$, we randomly sample from $D_\text{test}$ to match the class frequency of $D_\text{train}^i$ and distribute it to each client $i$. 
\end{itemize}


\begin{table}[t!]
    \centering
    \small
    \caption{Hyperparameters for VGG11 training on CIFAR-10.}
    \label{apptab:hyper_summary_10}
    \vspace{5pt}
    \resizebox{\textwidth}{!}{
    \begin{tabular}{c|ccccc|ccc}
      \toprule
      & \multicolumn{5}{c|}{Feature un-normalized algorithms} & \multicolumn{3}{c}{Feature normalized algorithms} \\ \midrule
      Hyperparameters & FedAvg & FedBABU & SCAFFOLD & FedNTD & FedExP & FedETF & SphereFed & \alg (Ours) \\ \hline
      $\eta$ & 0.01 & 0.01 & 0.01 & 0.01 & 0.01 & 0.05 & 0.55 & 0.35 \\ 
      Additional  & None & None & None & $(\beta, \tau)$=(1,3) & 
      $\epsilon$=0.001 & $(\beta, \tau)$=(1,1) & None & $\beta$=0.9 \\ \bottomrule
    \end{tabular}
    }
\end{table}

\begin{table}[t!]
    \centering
    \small
    \caption{Hyperparameters for MobileNet training on CIFAR-100.}
    \label{apptab:hyper_summary_100}
    \vspace{5pt}
    \resizebox{\textwidth}{!}{
    \begin{tabular}{c|ccccc|ccc}
      \toprule
      & \multicolumn{5}{c|}{Feature un-normalized algorithms} & \multicolumn{3}{c}{Feature normalized algorithms} \\ \midrule
      Hyperparameters & FedAvg & FedBABU & SCAFFOLD & FedNTD & FedExP & FedETF & SphereFed & \alg (Ours) \\ \hline
      $\eta$ & 0.1 & 0.1 & 0.1 & 0.1 & 0.1 & 0.5 & 6.5 & 5.0 \\ 
      Additional  & None & None & None & $(\beta, \tau)$=(1,3) & $\epsilon$=0.001 & $(\beta, \tau)$=(1,1) & None & $\beta$=0.9 \\ \bottomrule
    \end{tabular}
    }
\end{table}

\subsection{Hyperparameter Search for \texorpdfstring{$\eta$}{eta} and \texorpdfstring{$E$}{E}}\label{appsubsec:grid_search}

To optimize the initial learning rate ($\eta$) and the number of local epochs ($E$) for our algorithm, we conduct a grid search on the CIFAR-10 and CIFAR-100 datasets. The process and reasoning are outlined below.

\subsubsection{Rationale for Varying Initial Learning Rate \texorpdfstring{($\eta$)}{(eta)}}

The algorithms used in our experiments differ in handling feature normalization within the loss function. Some algorithms apply feature normalization, while others do not. When features $f(x;\bm{\theta})$ are normalized, the resulting gradient is scaled by $\frac{1}{\|f(x;\bm{\theta})\|_2}$. This scaling effect necessitates a grid search across various learning rates to account for the differences in learning behavior.

\subsubsection{Rationale for Varying Local Epochs \texorpdfstring{$E$}{E}}

In FL, choosing the appropriate number of local epochs is crucial. Too few epochs can lead to underfitting, while too many can cause client drift. Therefore, finding the optimal number of local epochs is essential by exploring a range of values.

\subsubsection{Grid Search Process and Results}

Considering the above reasons, we perform grid search for $\eta$ and $E$ on CIFAR-10 and CIFAR-100 datasets. The grid search for CIFAR-10 uses a shard size of 2, while for CIFAR-100, a shard size of 10 is used. The detailed procedures for each dataset are provided below. These optimal settings have also been confirmed to yield good performance in less heterogeneous settings.

\myparagraph{CIFAR-10.}
We examine $\eta$ values from \{0.01, 0.05, 0.1, 0.15, 0.2, 0.25, 0.3, 0.35, 0.4, 0.45, 0.5, 0.55, 0.6\}. For $E$, we consider \{1, 3, 5, 10, 15\}. The optimal learning rates vary by algorithm, and the results are summarized in \autoref{apptab:hyper_summary_10}. \autoref{apptab:hyper_summary_10} also includes the additional hyperparameters used for each algorithm. The notation for these additional hyperparameters follows the conventions used throughout this paper. The optimal number of local epochs is found to be 10 for every algorithm.

\myparagraph{CIFAR-100.}
We examine $\eta$ values from \{0.1, 0.3, 0.5, 1.0, 1.5, 2.0, 2.5, 3.0, 3.5, 4.0, 4.5, 5.0, 5.5, 6.0, 6.5, 7.0\}. A default initial learning rate of 0.1 is used unless specified otherwise. The optimal learning rates differ by algorithm, and the results are listed in \autoref{apptab:hyper_summary_100}. \autoref{apptab:hyper_summary_100} also includes the additional hyperparameters used for each algorithm. The notation for these additional hyperparameters follows the conventions used throughout this paper. The optimal number of local epochs is found to be 3 for every algorithm.

\newpage

\section{Additional Experiment Results}\label{appsec:add_exp}

\begin{table}[t!]
    \centering
    \caption{PFL accuracy comparison with MobileNet on CIFAR-100. For PFL, we denote the entries in the form of X{\tiny $\pm$(Y)}, representing the mean and standard deviation of personalized accuracies across all clients derived from a single seed.}
    \label{apptab:pfl_acc}
    \small
    \vspace{5pt}
    \resizebox{\textwidth}{!}{
    \begin{tabular}{l|ccc|ccc}
    \toprule
    Algorithm & $s$=10 & $s$=20 & $s$=100 & $\alpha$=0.05 & $\alpha$=0.1 & $\alpha$=0.3 \\ \midrule

    Dot-Regression                                   & 42.52  & 49.02 & 52.86  & 30.31{\tiny $\pm$7.95} & 37.52{\tiny $\pm$5.60} & 47.08{\tiny $\pm$3.69} \\
    
    Dot-Regression\,FT ($\mathcal{L}_\text{DR}$) & 80.84{\tiny $\pm$(5.99)}  & 74.18{\tiny $\pm$(5.78)} & 56.84{\tiny $\pm$(5.04)} & 72.02{\tiny $\pm$(6.80)}  & 66.96{\tiny $\pm$(5.36)} & 60.34{\tiny $\pm$(3.66)} \\ 
    {Dot-Regression\,FT ($\mathcal{L}_\text{Dr+}$)} & 80.82{\tiny $\pm$(6.12)}  & 73.73{\tiny $\pm$(5.75)} & 56.69{\tiny $\pm$(4.95)} & 71.85{\tiny $\pm$(7.03)}  & 66.59{\tiny $\pm$(5.32)} & 59.87{\tiny $\pm$(3.65)} \\ \midrule
    \alg (ours)                                   & 48.69  & 51.00 & 53.23   & 39.63{\tiny $\pm$9.12}  & 45.83{\tiny $\pm$6.18} & 48.04{\tiny $\pm$3.44} \\    
    \textbf{\alg\,FT ($\mathcal{L}_\text{DR}$) (ours)} & \textbf{84.23}{\tiny $\pm$(5.44)}  & \textbf{75.73}{\tiny $\pm$(4.79)} & \textbf{56.90}{\tiny $\pm$(4.85)} & \textbf{78.65}{\tiny $\pm$(6.17)}  & \textbf{74.86}{\tiny $\pm$(4.77)} & \textbf{62.47}{\tiny $\pm$(3.72)} \\

    \alg\,FT ($\mathcal{L}_\text{Dr+}$) (ours) & 84.10{\tiny $\pm$(5.43)}  & 75.42{\tiny $\pm$(4.80)} & 56.76{\tiny $\pm$(4.91)} & 78.55{\tiny $\pm$(6.16)}  & 74.75{\tiny $\pm$(4.75)} & 62.16{\tiny $\pm$(3.73)} \\ \bottomrule
    \end{tabular}
    }
\end{table}

\subsection{Personalized Federated Learning Results}
\label{subsec:pfl results}

We introduce \alg\,FT and dot-regression\,FT, inspired by prior work~\citep{oh2021fedbabu, dong2022spherefed, li2023no, kim2023fedfn}. These methods enhance personalization by leveraging local data to fine-tune the GFL model. We investigate the impact of fine-tuning using $\mathcal{L}_\text{Dr+}$ and $\mathcal{L}_\text{DR}$ loss for each GFL model to assess their effectiveness on personalized accuracy. Performance metrics without standard deviations indicate results on $D_\text{test}$, obtained from the GFL model after the initial step in the 2-step method. Our experiments involve heterogeneous settings with sharding and LDA non-IID environments, using MobileNet on CIFAR-100 datasets. We set $s$ as 10, 20, and 100, and the LDA concentration parameter ($\alpha$) as 0.05, 0.1, and 0.3. Table~\ref{apptab:pfl_acc} provides detailed personalized accuracy results.

Our 2-step process involves first developing the GFL model either using dot-regression or \alg. In the second step, we fine-tune this model to create the PFL model, again using $\mathcal{L}_\text{DR}$ or $\mathcal{L}_\text{Dr+}$. This results in four combinations: Dot-Regression\,FT\,($\mathcal{L}_\text{DR}$), Dot-Regression\,FT\,($\mathcal{L}_\text{Dr+}$), \alg\,FT\,($\mathcal{L}_\text{DR}$), and \alg\,FT\,($\mathcal{L}_\text{Dr+}$). When the GFL model is fixed, using $\mathcal{L}_\text{DR}$ for fine-tuning consistently outperforms $\mathcal{L}_\text{Dr+}$ across all settings, because dot-regression focuses on local alignment which advantages personalized fine-tuning. Conversely, when the fine-tuning method is fixed, employing $\mathcal{L}_\text{Dr+}$ for the GFL model consistently outperforms $\mathcal{L}_\text{DR}$ across all settings. This aligns with previous research~\cite {nguyen2022begin,chen2022importance} suggesting that fine-tuning from a well-initialized model yields better PFL performance.

\subsection{Elapsed Time Results}
\label{subsec:elapsed_time}

We compare \alg with various GFL algorithms for the elapsed time per communication round on CIFAR-100 ($s$=10). The results, detailed in \autoref{apptab:elapsed_time}, show that \alg exhibits a similar but slightly longer elapsed time than the other algorithms.

\begin{table}[h!]
    \centering
    \small
    \caption{Elapsed time per round (in seconds)  for various GFL algorithms.}
    \label{apptab:elapsed_time}
    \vspace{5pt}
    \addtolength{\tabcolsep}{-2pt}
    \resizebox{\textwidth}{!}{
    \begin{tabular}{c|ccccc|ccc}
      \toprule
      & \multicolumn{5}{c|}{Non-feature normalized algorithms} & \multicolumn{3}{c}{Feature normalized algorithms} \\ \midrule
     & FedAvg & FedBABU & SCAFFOLD & FedNTD & FedExP & FedETF & SphereFed & \alg (Ours) \\ \midrule
      Elapsed time  & 21.3 & 20.9 & 22.3 & 22.9 & 20.3 & 22.2 & 22.3 & \textbf{24.4} \\ \bottomrule
    \end{tabular}
    }
\end{table}



\end{document}